%% file: main.tex
\documentclass{article}
\usepackage{amssymb}
\usepackage{makecell}
\usepackage[final]{corl_2025} % Uncomment for the camera-ready ``final'' version.
\input{config.tex}

\title{\benchmark: Benchmarking Vision-Language Models for Low-Level Robot Manipulation}

\author{
  Enyu Zhao$^*$ \quad Vedant Raval$^*$ \quad Hejia Zhang\thanks{Denotes a co-lead author.} \quad \textbf{Jiageng Mao}\\
  \textbf{Zeyu Shangguan} \quad \textbf{Stefanos Nikolaidis} \quad \textbf{Yue Wang} \quad \textbf{Daniel Seita} \\
  Department of Computer Science, University of Southern California \\
}

\begin{document}
\doparttoc % Tell to minitoc to generate a toc for the parts
\faketableofcontents % Run a fake tableofcontents command for the partocs

\maketitle

%===============================================================================
\vspace{-8mm}
\begin{abstract}
\renewcommand*{\thefootnote}{\arabic{footnote}}
Vision-Language Models (VLMs) have revolutionized artificial intelligence and robotics due to their commonsense reasoning capabilities. In robotic manipulation, VLMs are used primarily as high-level planners, but recent work has also studied their lower-level reasoning ability, which refers to making decisions about precise robot movements. However, the community currently lacks a clear and common benchmark that can evaluate how well VLMs can aid low-level reasoning in robotics. Consequently, we propose a novel benchmark, \benchmark, to evaluate the low-level robot manipulation reasoning capabilities of VLMs across various dimensions, including how well they understand object-object interactions and deformable object manipulation. We extensively test 33 representative VLMs across 10 model families on our benchmark, including variants to test different model sizes. Our evaluation shows that the performance of VLMs significantly varies across tasks, and there is a strong correlation between this performance and trends in our real-world manipulation tasks. It also shows that there remains a significant gap between these models and human-level understanding. 
%Our benchmark’s questions and code will be made open-source. 
See our website at: \href{https://manipbench.github.io}{https://manipbench.github.io}.
\end{abstract}


%===============================================================================

\section{Introduction}

One long-standing goal in robotics is to train a ``generalist" robot capable of performing diverse tasks, particularly robot manipulation. A promising paradigm for this is to leverage the broad knowledge in Vision-Language Models (VLMs) such as GPT-4~\cite{GPT-4} and Gemini~\cite{Gemini}. While the community has used VLMs to achieve great generalization in domains like computer vision and natural language processing, robotics faces unique challenges with requiring either difficult-to-scale physical real-world interaction data or simulation data with sim-to-real gaps, making it challenging for VLMs to act as low-level planners. However, recent work has extensively explored incorporating these ``foundation" models~\cite{Bommasani2021FoundationModels} such that they can generate low-level trajectories executable by an embodiment~\citep{LLMs_zero_shot_traj, GPTFabric, MOKA2024, wangwonderful}. This direction is especially important because it offers a path to bypass large-scale, task-specific data collection by leveraging general-purpose pre-trained models. Beyond improving scalability, this enables faster deployment in open-world settings where generalization to unseen tasks and objects is critical. 
It remains unclear, however, which is the optimal foundation model for a ``VLM agent'' in tasks like fabric or articulated object manipulation, and how VLMs perform in low-level reasoning tasks required for manipulation. 

\begin{figure*}[t]
\center
\includegraphics[width=1.0\textwidth]{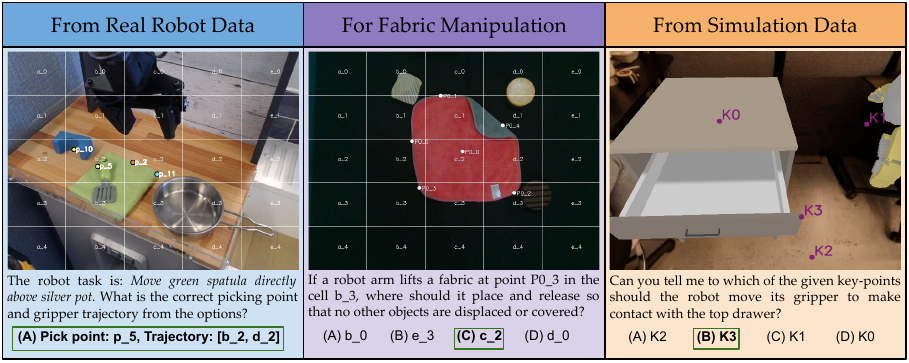}
\caption{\benchmark is a novel benchmark with over 12,000 multiple-choice questions across three different categories to evaluate low-level physical reasoning capabilities of VLMs in the context of robotic manipulation. For the first question (from real robot data), we do not display the text of all options due to space limitations.}
\label{fig:pull}
\vspace*{-10pt}
\end{figure*}

Motivated from these questions, we propose \benchmark: a novel open-source benchmark to evaluate how well VLMs understand the low-level effect of a robot's action on its environment (see Fig.~\ref{fig:pull}). While there exist benchmarks to evaluate VLMs for robotics~\citep{wang2023newton,PhysBench2025,Octopi2024,zhang2024vlabenchlargescalebenchmarklanguageconditioned,gao2024physicalVLM,guruprasad2024benchmarkingvisionlanguage,10802733,fu2024can,garcia24gembench}, our approach and benchmark differ significantly along axes such as task diversity, model diversity, and particularly our novel multiple-choice question (MCQ) based evaluation design, which efficiently assesses the low-level reasoning capabilities of VLMs without requiring trajectory rollouts, as detailed in Table \ref{tab:pull}. We evaluate 33 VLMs across 10 families (2 closed-source and 8 open-source), including size-based variants, to assess their strengths and weaknesses. The best models, such as Gemini-2.5-pro, significantly outperform random chance and other models on multiple choice questions, but still show substantial room for improvement, highlighting the need for further innovation in VLM development. Furthermore, we include experiments indicating a significant connection between performance on our benchmark versus performance in the real world when using VLMs to select robot actions, validating the consistency of \benchmark. 
In summary, this paper contributes:
\begin{enumerate}[leftmargin=*,noitemsep]
    \item \benchmark, an MCQ-based benchmark for evaluating VLMs’ reasoning for low-level robotic manipulation, consisting of \textbf{\QuestionNum} questions across tasks ranging from pick-and-place, articulated object manipulation, deformable object manipulation, and dynamic manipulation. %and tool-use manipulation tasks.
    \item Extensive evaluation of various VLM families to assess different dimensions of reasoning.
    \item Real world experiments on separate manipulation tasks, demonstrating a significant correlation between a model's \benchmark performance and its effectiveness at selecting robot actions. % Daniel: my concern with saying ``held-out'' is that it suggests we have a set of N tasks and picked 80% for train, 20% for test, like a machine learning model. I slightly edited this to say ``separate'' -- maybe we can put in the unseen/held-out explanation somewhere in the text itself (not in the bullet point list).
\end{enumerate}

%===============================================================================

\section{Related Work}
\label{sec:rel_work}

% \subsection{Vision-Language Models in Robotics}
\textbf{Vision-Language Models in Robotics.} 
VLMs have increasingly become effective as high-level planners~\citep{ahn2022icanisay,driess2023palme,huang2022language} which can produce executable robot code~\citep{codeaspolicies,progprompt,huang2023voxposer}. 
However, VLMs can have difficulty reasoning about spatial and physical properties~\cite{PhysBench2025}, which are essential for manipulation~\cite{manipulation}.
This has led to work on improving their zero-shot spatial reasoning via methods such as iterative prompting~\cite{PIVOT} and annotating images~\cite{yang2023setofmark}. Another line of work focuses on fine-tuning VLMs to improve spatial reasoning~\citep{SpatialVLM,gao2024physicalVLM,tangandrajkumar2024kalie,RoboPoint_2024}. 
In contrast, our objective is to systematically evaluate VLMs for predicting robot actions and to identify which VLMs are best as ``agents'' for a robot. 
We inspect the affordance reasoning capabilities of VLMs using the MOKA framework~\citep{MOKA2024} which leverages VLMs to predict keypoints~\cite{manuelli2019isrr} (\ie affordances) to define robot actions.  %and inspect the affordance reasoning capabilities of VLMs. 
VLMs are also closely related to robotic foundation models~\cite{FMs_robotics_2023_survey_1,FMs_robotics_2023_survey_2}, also referred to as Vision-Language-Action (VLA) models. Some examples of these include RT-2~\cite{rt22023arxiv}, Octo~\cite{octo_2024}, OpenVLA~\cite{kim24openvla}, and $\pi_0$~\cite{black2024pi0visionlanguageactionflowmodel,pertsch2025fastefficientactiontokenization}. Our work is complementary to these, as we focus on benchmarking VLMs which are not robotics-specific. In addition, evaluating general VLMs can help understand their impact on robotic foundation models if they are part of them (e.g., OpenVLA uses Llama 2 7B~\cite{touvron2023llama}).  % Daniel: slightly reword later, 'if they are part of them' sounds awkward.

\begin{table}[t]
    \centering
    \renewcommand{\arraystretch}{1.2}
    \setlength{\tabcolsep}{6pt}
    \scriptsize
    \resizebox{\linewidth}{!}{
    \begin{tabular}{c c c c c c c c c }
        \toprule
        & \makecell{Low-level \\ Manip. Reasoning} & \makecell{Manip. Task \\ Factorization} & \makecell{Perf. Trans. \\ Eval.} & \makecell{Real World \\ Data} & \makecell{Fabric \\ Manipulation}  & \makecell{Dynamic \\ Manipulation} & \makecell{Model \\ Diversity}  \\
        \midrule
        PhysBench~\citep{PhysBench2025} & \xmark & \xmark & \cmark & \cmark & \xmark & \xmark & High \\
        NEWTON~\citep{wang2023newton} & \xmark & \xmark & \xmark & \xmark & \xmark & \xmark & Medium \\ 
        VLABench~\citep{zhang2024vlabenchlargescalebenchmarklanguageconditioned} & \cmark & \cmark & \cmark & \xmark & \xmark & \xmark & Low \\
        PhysObjects*~\citep{gao2024physicalVLM} & \xmark & \xmark & N/A & \cmark  & \xmark & \xmark & N/A \\
        % The Colosseum~\citep{Pumacay-RSS-24} \\
        MultiNet~\citep{guruprasad2024benchmarkingvisionlanguage} & \cmark & \xmark & \xmark & \cmark & \cmark & \cmark  & Low  \\
        % FactorWorld~\citep{10611331} \\
        % VLATest~\citep{wang2024towards} & \cmark & \cmark & \xmark & \xmark & \xmark & \xmark & Medium \\
        Octopi*~\citep{Octopi2024} & \xmark & \xmark & N/A & \cmark & \xmark & \xmark & N/A \\
        Open6DOR~\citep{10802733} & \cmark & \cmark & \xmark & \cmark & \xmark & \xmark & Medium \\
        ETPBench~\citep{fu2024can} & \xmark & \cmark & \xmark & \xmark & \xmark & \xmark & Medium \\ 
        GemBench*~\citep{garcia24gembench} & \cmark & \xmark & N/A & \xmark & \xmark & \xmark & N/A \\
        \midrule
        \benchmark & \cmark & \cmark & \cmark & \cmark  & \cmark & \cmark & High \\
        \bottomrule
    \end{tabular}
    }
    \vspace{4pt}
   %\makebox[\linewidth][l]{\footnotesize * denotes datasets.}
    %\vspace{2pt}
    \caption{Comparison of \benchmark and other vision-language benchmarks or datasets (denoted with *). \benchmark is the only vision-language benchmark that evaluates low-level manipulation reasoning in VLMs using multiple-choice questions, enabling efficient and effective assessment. See Appendix~\ref{appendix:table} for more details.}
    \label{tab:pull}
    \vspace{-15pt}
\end{table}

%\subsection{Benchmarking LLMs and VLMs}
\textbf{Benchmarking LLMs and VLMs.} 
Alongside the rapid advances in LLMs and VLMs, significant work has focused on benchmarking these models. Popular benchmarks evaluate mathematical reasoning~\citep{hendrycksmath2021,gao2024omnimathuniversalolympiadlevel}, trust and safety~\citep{huang2024trustllm}, and visual tasks like interpreting charts and maps~\citep{SEED-Bench-2-2024}. Others, such as AgentBench~\citep{AgentBench2024}, assess LLMs as agents in code, game, and web environments. In contrast, our benchmark centers on robotics, complementing existing efforts in VLM-based mapping and navigation~\citep{MANGO2024}, compositional reasoning~\citep{zheng2022vlmbench}, and task planning for embodied AI~\citep{choi2024lotabench,li2024embodiedagentinterfacebenchmarking,zhang2024etplanbenchembodiedtasklevelplanning}.

%\vedant{Modify this paragraph based on Table~\ref{tab:pull}} 
More closely-related benchmarks, such as NEWTON~\citep{wang2023newton} and Octopi~\citep{Octopi2024}, evaluate physical reasoning capabilities. Here, VLMs answer multiple choice questions about object properties (such as whether an object is ``brittle’’ or ``soft'') from language and, for Octopi, high-resolution tactile data. 
Recently, the PhysBench~\cite{PhysBench2025} benchmark evaluates VLMs using multiple choice questions to test spatial reasoning capabilities. 
In contrast, \benchmark does not involve explicitly predicting object properties, and it evaluates how well VLMs can directly predict keypoints which define a low-level action for a robot. 
%% Furthermore, while we similarly evaluate based on multiple choice questions, \benchmark focuses on low-level action prediction in robotics, instead of spatial reasoning. 
%We furthermore evaluate VLMs and their capabilities to reason about different physics properties (\eg object-object interaction).  See Section~\ref{sec:benchmark} for a further discussion of ``low-level'' and ``high-level'' actions.
Other benchmarks for VLMs and robot manipulation include MultiNet~\cite{guruprasad2024benchmarkingvisionlanguage} and VLABench~\cite{zhang2024vlabenchlargescalebenchmarklanguageconditioned}. 
MultiNet uses data from Open-X~\cite{open_x_embodiment_rt_x_2023} and assesses how well VLMs predict trajectories using mean square error (MSE). However, using MSE for evaluation may not adequately measure performance in the case of multimodality. 
VLABench evaluates VLMs on manipulation tasks and assumes the presence of a skill library, whereas we do not use a skill library. 
In contrast to prior benchmarks, \benchmark also has a much greater focus on the important topic of deformable object manipulation~\cite{manip_deformable_survey_2018,2021_survey_defs}. 
See Table~\ref{tab:pull} for an overview comparison. 
%See the Appendix for an extended discussion for how \benchmark differentiates from recent and concurrent benchmarks. % Daniel (Jan 05): I think we need to have this because I think there's so many benchmarks in the last few months that this differentiation will be important. 

% \subsection{Benchmarks and Datasets in Robot Manipulation}
\textbf{Benchmarks and Datasets in Robot Manipulation.} 
Robotics benchmarks and simulation environments are critical to evaluate algorithms and to measure progress in robot manipulation. 
% To keep our scope manageable, we focus on those for robot manipulation~\citep{manipulation}. 
Benchmarks for high-level planning with mobile manipulators include BEHAVIOR-1K~\citep{BEHAVIOR1K}, AI2-THOR~\citep{ai2thor}, Habitat 2.0~\citep{szot2021habitat}, and RoboCasa~\citep{robocasa2024}. 
Our main objective, however, is to study how well VLMs understand lower-level and more precise manipulation, though we could still leverage such simulators for data collection. 
Other manipulation benchmarks focus on low-level control using high-DOF hands~\citep{robopianist2023} or humanoids~\citep{chernyadev2024bigym,sferrazza2024humanoidbench}; these are complementary and out of scope. 
Benchmarks closer in scope to \benchmark test deformable object manipulation~\citep{UBSoft_2024,corl2020softgym,seita_bags_2021}. \benchmark can leverage these to create questions for deformable manipulation reasoning capabilities. 

More general manipulation task suites include MetaWorld~\cite{yu2019metaworld}, Ravens~\citep{zeng_transporters_2020}, CALVIN~\cite{mees2022calvin}, RoboSuite~\citep{robosuite2020}, ManiSkill~\citep{gu2023maniskill2,tao2024maniskill3gpuparallelizedrobotics}, and RLBench~\citep{RLBench_2020}.  
These study aspects of robot manipulation and propose new simulation tasks and environments. Our benchmark is complementary; \benchmark contains questions to evaluate VLMs. Thus, as the community creates more task-related robotics benchmarks, these provide an expanding source of data for \benchmark. 
Furthermore, our benchmark uses large-scale data collected from the robotics community, including DROID~\citep{khazatsky2024droid} and Bridge~\citep{walke2023bridgedata} from Open-X~\citep{open_x_embodiment_rt_x_2023}. 
Thus, much of our benchmark's data is already used in practice.

%===============================================================================

\section{\benchmark: Overview}
\label{sec:benchmark}

\begin{figure*}[t]
\center
\includegraphics[width=1.0\textwidth]{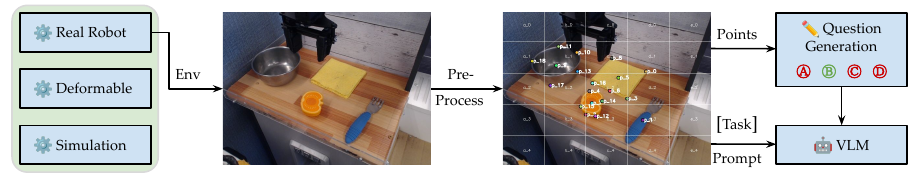} 
\caption{
\benchmark uses real and simulated environments, typically pre-processed with a MOKA-style~\cite{MOKA2024} pipeline to extract key-points and grid annotations for generating VLM-evaluable multiple choice questions.}
\label{fig:systems}
\vspace*{-10pt}
\end{figure*}

\benchmark comprises \textbf{\QuestionNum} multiple-choice questions (MCQs) spanning diverse domains. These questions are categorized based on their origin (see Sec.~\ref{section:data}): those derived from existing real-world datasets, those manually curated by us for fabric manipulation, and those sourced from simulation data. 
Our preliminary experiments reveal that questions centered on robot action trajectories and the keypoints guiding those trajectories provide the most valuable insights when evaluating VLMs. Thus, we use mark-based visual prompting to curate the MCQs, which primarily focus on selecting appropriate \textit{interaction keypoints}: contact-initiation (e.g., picking points), contact-release (e.g., placing points), and post-contact motion (e.g., pushing direction points). %\hejia{want to make the definition of keypoints more generalizable than just picking and placing} 
To achieve this, we process the collected data from each origin. The data, mainly consisting of the image observations (with some exceptions), is used to formulate the multiple-choice questions. The necessary information, such as the action trajectories, is used to guide the option generation, along with acting as the ground-truth for evaluating the VLMs. We incorporate a MOKA-style~\cite{MOKA2024} pipeline for most of the image processing involved in \benchmark. See Fig.~\ref{fig:systems} for an overview.

%===============================================================================

\section{\benchmark: Data Preparation} 
\label{section:data}

%\hejia{Primarily discuss how we collect raw image observations from different sources in this section.}
%We leverage three distinct data sources to collect robot sequential image observations in different tasks and environments, which we subsequently use to generate evaluation questions (Fig.~\ref{fig:pull}). 

\textbf{From public robotic manipulation datasets.} We use demonstrations from the open-source robot manipulation datasets of DROID~\cite{khazatsky2024droid} and Bridge~\cite{walke2023bridgedata} for curating \benchmark questions. 
We study these datasets because of their importance in large-scale robotic imitation learning~\cite{open_x_embodiment_rt_x_2023}. 
We separate the DROID data into two subsets based on the manipulation task: \emph{DROID articulated (art.)} and \emph{DROID pick-and-place (p\&p)}. 
To prepare the data and generate questions, we extract the ground-truth gripper trajectories to obtain the picking and placing points by finetuning GroundingDINO~\citep{liu2024grounding} and leveraging the data from~\citep{kuang2024ramretrievalbasedaffordancetransfer}, which contains human-annotated gripper positions for affordance prediction across multiple datasets. See Appendix~\ref{appendix:data_preparation_public} for details.

% \hejia{I commented out the second paragraph, since it's talking about processing raw data, which is repeated in Sec. 5}

% Vedant (04/20) - Much of this information is already getting repeated in the Appendix. Hence commenting this.
% A key bottleneck here is detecting the gripper on the 2D image, since this information is not directly available. To this end, we use the data from~\citep{kuang2024ramretrievalbasedaffordancetransfer}, which contains human-annotated gripper positions for affordance prediction across multiple datasets. However, the annotations are not always complete. To detect the entire trajectory, we use their annotated subset of the DROID and Bridge data and incorporate Grounding DINO~\citep{liu2024grounding} to track the gripper throughout the different episodes to efficiently generate image-mask pairs. See the Appendix for more details. 

\textbf{From in-house fabric manipulation setup.} Given its practical importance~\cite{black2024pi0visionlanguageactionflowmodel,grasp_centered_survey_2019,longhini2024unfoldingliteraturereviewrobotic}, we manually curate data specifically designed to evaluate VLMs' understanding of fabric manipulation, such as folding or smoothing tasks. 
We break down common aspects of fabric manipulation into ten distinct dimensions, such as understanding of fabric-object interactions and understanding of inverse dynamics. Each dimension represents a fundamental aspect that an agent must implicitly grasp to successfully perform fabric manipulation.
We get data from our real-world workstation. This comprises of a {3cm thick 98cm $\times$ 78cm} foam, an Intel Realsense d415i RGBD camera mounted at a height of 87.2cm, rectangular fabrics, and solid objects of varying dimensions. This setup is used to capture top-down image observations of numerous scenes in different task settings for formulating questions (Sec.~\ref{sec:q_gen_fabric}). The data collection procedure varies slightly across different dimensions. See Appendix~\ref{appendix:data_preparation_fabric} for the description, choice, and importance of the dimensions.
%\hejia{I commented out the sentence talks about noting the necessary information, since it's confusing how it will be used later.}
% We manually note the necessary information, such as the desired fabric key-point(s), for formulating the questions. 

\textbf{From simulation}. To assess manipulation tasks where real-world deployment may be cumbersome, such as tool use and dynamic manipulation, we compile a suite of simulated tasks that spans all task categories in \benchmark: (i) pick-and-place, (ii) articulated object manipulation, (iii) deformable object manipulation, (iv) tool manipulation, and (v) dynamic manipulation. We primarily adapt simulation assets, and pre-trained policies from popular existing benchmarks: SimplerEnv~\citep{li24simpler}, RLBench~\citep{RLBench_2020}, and SoftGym~\citep{corl2020softgym} to generate data for our evaluation questions (Sec.~\ref{sec:q_gen_sim}). We use environments from SimplerEnv for tasks (i) and (ii), while using SoftGym and RLBench, respectively, for tasks (iii) and (iv). For task (v), we construct a new ball-shooting environment in IsaacSim~\cite{isaacsim}. Additional details on data collection for each task are presented in Appendix~\ref{appendix:data_preparation_sim}. % (\hejia{specific section}).

%===============================================================================

\section{\benchmark: Question Generation}
\label{sec:gen_questions}

Using the data sources described in Sec.~\ref{section:data}, we generate \textbf{\QuestionNum} multiple-choice questions. We present preprocessing steps (Sec.~\ref{ssec:preprocessing}) and how we generate the three types of MCQs (Sec.~\ref{sec:q_gen_droid},~\ref{sec:q_gen_fabric}, and~\ref{sec:q_gen_sim}).

\subsection{Preprocessing Steps for Generating MCQs on Real World Data}
\label{ssec:preprocessing}
\vspace{-1.5pt}

%Our preliminary experiments reveal that questions centered on robot action trajectories and the keypoints guiding those trajectories provide the most valuable insights when evaluating VLMs. Therefore, our MCQs primarily focus on prompting VLMs to select appropriate \textit{interaction keypoints}: contact-initiation (e.g., picking points), contact-release (e.g., placing points), and post-contact motion (e.g., pushing direction points) for different manipulation tasks.

% To construct such keypoint selection questions, we first pre-process our collected robot sequential image observations (Sec.~\ref{section:data}) to annotate candidate \textit{interaction keypoints} for selection. We use slightly different pre-processing approaches for data from different sources. 

% \hejia{I feel this section is messed up, in 5.1 paragraph 2, we mentioned the ground-truth key-point. In 5.2 we mentioned we detect the gripper to obtain ground truth points again. When did we get ground truth points?}

%\noindent\textbf{Preprocessing for real world data.} 
For MCQs based on real world data, we begin by pre-processing image observations following MOKA~\cite{MOKA2024}. Given an observation image and a natural language task description, we prompt GPT-4o to identify the key object $O_K$ for the task. We use Grounded SAM~\cite{ren2024grounded} to segment all objects ($O_K$ and any others) and obtain their image masks. From each object mask, we sample one point from the center and the rest from its contour using farthest-point sampling~\cite{moenning2003fastfps}. We then annotate the original observation image with the sampled points, and a grid overlay (usually $5\times5$). This annotated image encodes the necessary information of objects and environments, and will serve as the image prompt for VLMs (e.g., see Fig.~\ref{fig:pull}, first two examples).

We use the prepared ground-truth keypoint $p_{g}$ (e.g., pick and place point) to generate the correct choice for the multiple-choice questions. To ensure that the correct picking point lies on the key object $O_K$, we replace the ground-truth picking point $p_{pick}$ with the closest point among those sampled from $O_K$, if the distance between those two points is lower than a threshold. The process to generate the natural language prompts for the VLMs, along with the incorrect options, is different for the three question types. To assist VLMs, we include CoT prompting~\citep{wei2022chainofthought} in the questions. 

% \noindent\textbf{Preprocessing for simulation data.} For tasks in simulation, we leverage the proprietary information from simulators to annotate candidate keypoints for selection. 

\subsection{MCQs from Public Robotic Manipulation Datasets}\label{sec:q_gen_droid}
\vspace{-1.5pt}

For each episode, we identify the frames $f_s$ and $f_e$ where the manipulation ``starts'' and ``ends,'' respectively. For each frame, we follow the pre-processing technique as described in Sec.~\ref{ssec:preprocessing} to construct vision-language question prompts for VLMs. We design two types of questions:

\textbf{Type 1 (Q1)}:  Given an image and language description of a task, the VLM selects the best matching trajectory from four candidates. Each trajectory includes a picking keypoint and two image tiles: the start tile (containing $p_{pick}$) and the end tile (containing $p_{place}$).

\textbf{Type 2 (Q2)}: We derive type 2 questions from type 1 by having the VLM first choose a picking point from four candidates. Then, it selects the appropriate ending tile using the ground-truth picking point. Both are sourced from the type 1 candidate trajectories.

Overall, 
% across {612} tasks, 
we have 9180 questions with 6120 type 1 questions and 3060 type 2 questions. Pure random guessing on type 1 questions will lead to a success rate of 25\% since there are always 4 choices, while random guessing for type 2 questions will be worse than 25\% performance since it combines predictions from picking and placing. We perform ``question augmentation" to build several questions from the same episode. See Appendix~\ref{appendix:question_generation_public} for more details.

\subsection{MCQs from In-house Fabric Manipulation Setup}
\label{sec:q_gen_fabric}
\vspace{-1.5pt}

We pre-process all the image observations following Sec.~\ref{ssec:preprocessing}. We discard data where the preprocessing pipeline failed (e.g., Grounded SAM could not detect keypoints). With the desired keypoints and grids, we manually create choices for the multiple-choice questions. We generate 2662 questions across all the ten dimensions. We do not perform question augmentation, as done for the other question categories to encourage diverse scenarios being considered, given the simplicity of many dimensions. Thus, in general, we only form one question for each recorded scene. The questions vary across the dimensions, based on what they are trying to evaluate. Pure random guessing will result in 25\% success given 4 choices.  Additional details and sample questions are in Appendix~\ref{appendix:question_generation_fabric}.

\subsection{MCQs from Simulation}
\label{sec:q_gen_sim}
\vspace{-1.5pt}

\begin{figure*}[t]
\begin{center}
\includegraphics[width=1.0\textwidth]{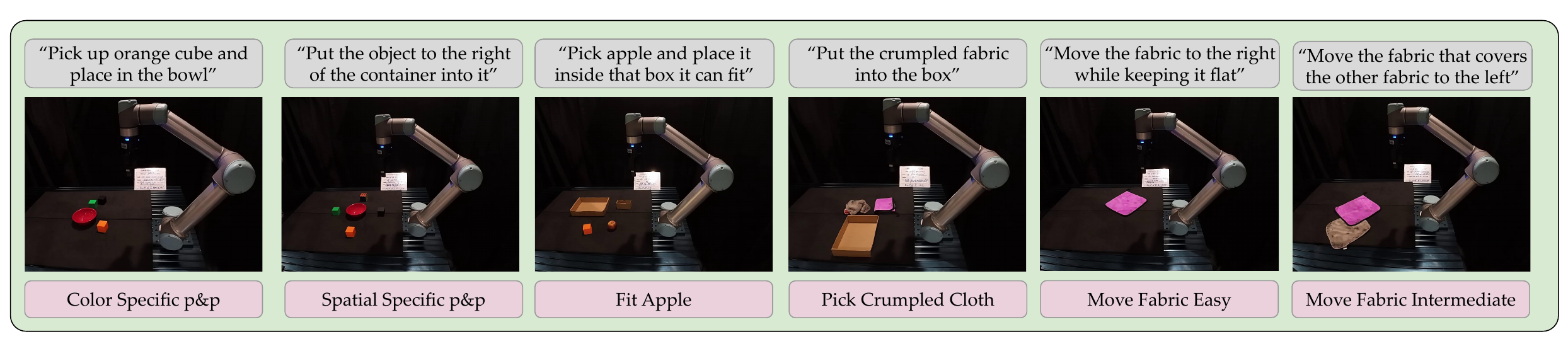}
\end{center}
\vspace{-8pt}
\caption{
    The different tasks, with their descriptions and initial states, present in our real-world experiments.
}
\label{fig:result_real_world}
\vspace{-10pt}
\end{figure*}

For the questions generated from existing simulation environments, we do not need to perform pre-processing steps (from Sec.~\ref{ssec:preprocessing}) given the access to the ground-truth keypoint information. In this case, we plot the ground-truth point, along with sampling some incorrect points, on the image (Fig.~\ref{fig:pull}, the third example). We do not overlay a grid pattern for these questions. We consider candidate object contact points and after-contact movements on the frames to generate evaluation questions.
VLMs are required to select one of four candidate keypoints to complete the given manipulation tasks. 
Pure random guessing will result in 25\% success for all tasks. See Appendix~\ref{appendix:question_generation_sim} for details.

%===============================================================================

\section{\benchmark: Evaluation Plan}
\label{sec:evaluation}

We evaluate 10 VLM families on \benchmark (see Table~\ref{tab:num-models} for details). This includes GPT and Gemini, two popular industry-backed closed-source model families widely regarded as among the most effective and versatile models in the Generative AI community. The other models are open-source and are known for their effective performance in multi-modal tasks (e.g., InternVL~\cite{chen2023internvl} and Qwen-VL~\cite{Qwen-VL}). %We consider different variants of these models. 
These evaluations were conducted mainly on 2 RTX 4090 GPUs. Another server with 5 RTX 6000 Ada GPUs was used for larger models.
Our evaluation metric is the percentage accuracy of correctly answering the questions. For each model and question type, we average over all the questions to report the numbers. 
As an upper bound baseline to compare with human intuition, we have 36 human volunteers evaluate these questions, through a custom web portal. The volunteers answer a subset of 50-70 questions for each question type, and we use their accuracies to obtain the final human score. We provide more details on the human evaluation in Appendix~\ref{appendix:human_evaluation}.

%\subsection{Real World Robot Manipulation}
\textbf{Real-World Robot Manipulation.} 
%In addition to evaluating VLMs on curated MCQs in \benchmark, 
We also conduct physical robot manipulation experiments to verify whether performance on \benchmark translates to unseen robot tasks in real world. 
We design 7 manipulation tasks whose specific combinations of task goals, language instructions, camera angles, and objects never appear together in \benchmark: \emph{Color Specific p\&p, Spatial Specific p\&p, Fit Apple, Pick Crumpled Cloth, Move Fabric Easy, Move Fabric Intermediate}, and \emph{Move Fabric Hard} (Fig.~\ref{fig:result_real_world})\footnote{The starting states for \emph{Move Fabric Easy} and \emph{Move Fabric Hard} are the same, so we only show one.}. 
In these experiments, VLMs select keypoints to define robot actions. The pipeline for generating candidate keypoints for real world tasks is similar to the one for generating MCQs in \benchmark (Sec.~\ref{ssec:preprocessing}).
We use a UR5 robot arm with a Robotiq 2F-85 parallel-jaw gripper and a top-down RGBD camera.
%. \vedant{Add information on the physical setup for the real-world experiments}.  
See Appendix~\ref{appendix:real_experiments} for more details about the real world experiments.
% See Fig~\ref{fig:result_real_world} for the real-world experiment pipeline.

%===============================================================================

\section{\benchmark: Results, Analysis, and Key Insights}
\label{sec:results}

\begin{table*}[t]
\centering

% First table: Bridge + DROID
\begin{minipage}{0.505\linewidth}
    \centering
    \setlength{\tabcolsep}{4pt}
    \scriptsize
    \resizebox{\linewidth}{!}{
    \begin{tabular}{lcccccc}
    \toprule
    \makecell{\bf Model} & \multicolumn{2}{c}{\bf Bridge} & \multicolumn{2}{c}{\bf DROID (art.)} & \multicolumn{2}{c}{\bf DROID (p\&p)} \\
    & \bf Q1 & \bf Q2 & \bf Q1 & \bf Q2 & \bf Q1 & \bf Q2 \\
    \midrule
    \textbf{Closed-Source} & & & & & & \\
    o1                     &\third{0.866} & 0.458 & 0.805 & 0.384 & \third{0.822} & \first{0.542} \\
    GPT-4.1                & 0.656 & 0.401 & 0.683 &  0.301  & 0.724 & 0.433   \\
    GPT-4o                 & 0.503 & 0.309 & 0.687 & \third{0.396} & 0.676 & 0.401 \\
    GPT-4o-mini            & 0.487 & 0.183 & 0.524 & 0.156 & 0.572 & 0.210 \\
    Gemini-2.5-pro         & \first{0.916} & 0.403    & \first{0.909} & 0.362    & \first{0.869}  &   0.453   \\
    % Gemini-2.0-flash-think* & ~     &  ~    & ~     &  ~    & ~     &  ~    \\
    Gemini-2.0-flash       & 0.594 & 0.287 & 0.645 & 0.387 & 0.560 & 0.235 \\
    Gemini-1.5-pro         & 0.398 & 0.309 & 0.414 & 0.126 & 0.384 & 0.158 \\
    Gemini-1.5-flash       & 0.458 & 0.347 & 0.378 & 0.123 & 0.324 & 0.193 \\
    \midrule
    \textbf{Open-Source} & & & & & & \\
    GLM-4V-9B               & 0.593 & 0.249 & 0.481 & 0.381 & 0.463 & 0.292 \\
    
    InternVL2-1B            & 0.015 & 0.028 & 0.000 & 0.105 & 0.000 & 0.066 \\
    InternVL2-2B            & 0.330 & 0.052 & 0.139 & 0.120 & 0.197 & 0.052 \\
    InternVL2-4B            & 0.341 & 0.134 & 0.261 & 0.283 & 0.317 & 0.242 \\
    InternVL2-8B            & 0.445 & 0.258 & 0.282 & 0.228 & 0.406 & 0.258 \\
    InternVL2-26B           & 0.634 & 0.326 & 0.545 & 0.293 & 0.593 & 0.296 \\
    InternVL2-40B           & 0.694 & 0.338 & 0.656 & 0.272 & 0.654 & 0.313 \\
    InternVL2-76B           & 0.748 & \third{0.509} & 0.601 & 0.235 & 0.632 & 0.411 \\

    InternVL2.5-1B          & 0.243 & 0.016 & 0.213 & 0.099 & 0.221 & 0.057 \\
    InternVL2.5-2B          & 0.404 & 0.026 & 0.182 & 0.096 & 0.244 & 0.027 \\
    InternVL2.5-4B          & 0.639 & 0.199 & 0.326 & 0.181 & 0.559 & 0.283 \\
    InternVL2.5-8B          & 0.600 & 0.330 & 0.348 & 0.229 & 0.511 & 0.272 \\
    InternVL2.5-26B         & 0.780 & 0.376 & 0.677 & \first{0.433} & 0.736 & 0.389 \\
    InternVL2.5-38B         & \second{0.904} & \second{0.528} & \second{0.829} & 0.316 & \second{0.839} & 0.425 \\
    InternVL2.5-78B         & 0.851 & \first{0.541} & 0.745 & 0.348 & 0.722 & 0.431 \\
    
    QwenVL-Chat             & 0.224 & 0.067 & 0.220 & 0.141 & 0.292 & 0.077 \\
    Qwen2VL-2B              & 0.237 & 0.045 & 0.223 & 0.156 & 0.204 & 0.089 \\
    Qwen2VL-7B              & 0.479 & 0.191 & 0.375 & 0.366 & 0.536 & 0.329 \\
    Qwen2VL-72B             & 0.670 & 0.460 & 0.645 & 0.395 & 0.742 & \third{0.460} \\

    Qwen2.5-VL-3B           & 0.428 & 0.190 & 0.335 & 0.144 & 0.512 & 0.197 \\
    Qwen2.5-VL-7B           & 0.548 & 0.344 & 0.430 & 0.372 & 0.602 & 0.408 \\
    Qwen2.5-VL-32B          & 0.649 & 0.428 & 0.632 & \second{0.399} & 0.574 & 0.459 \\
    Qwen2.5-VL-72B          & 0.809 & 0.470 &\third{0.809} & 0.390 & 0.796 & \second{0.481} \\

    LLaVA-NeXT-7B           & 0.228 & 0.071 & 0.111 & 0.094 & 0.206 & 0.100 \\
    Llama3.2-11B-VI         & 0.264 & 0.101 & 0.292 & 0.115 & 0.242 & 0.111 \\
    \midrule
    Random & 0.250 & 0.061 & 0.250 & 0.063 & 0.250 & 0.084 \\
    Human & 0.880 & 0.825 & 0.990 & 0.940 & 0.980 & 0.635 \\
    \bottomrule
    \end{tabular}
        }
    \caption{Performance comparison of various VLMs on our MCQs for the Bridge and DROID datasets. Each dataset includes two question types detailed in Sec.~\ref{sec:q_gen_droid}. %\enyu{* means this model's result is not there yet.}
    }
    \label{tab:bridge_droid_results}
\end{minipage}%
\hfill
% Second table: Simulation Tasks
\begin{minipage}{0.477\linewidth}
    \centering
    \setlength{\tabcolsep}{3pt}
    \scriptsize
    \resizebox{\linewidth}{!}{
    \begin{tabular}{lccccc}
    \toprule
    %\makecell{\bf Model} & \makecell{\bf Place \\ \bf Carrot} & \makecell{\bf Close \\ \bf Drawer} & \makecell{\bf Straight. \\ \bf Rope} & \makecell{\bf Sweep \\ \bf Object} & \makecell{\bf Ball \\ \bf Shoot.} \\
    \makecell{\bf Model} & {\bf Place} & \bf Close & \bf Straight. & \bf Sweep & \bf Ball \\
          & {\bf Carrot} & \bf Drawer & \bf Rope & \bf Object & \bf Shoot. \\
    \midrule
    \textbf{Closed-Source} & & & & & \\
    o1 & 0.776 & \third{0.627} & 0.721 & 0.608 & \third{0.593} \\
    GPT-4.1 & \second{0.798} & \second{0.687} & \first{0.800} & 0.644 & 0.506 \\
    GPT-4o & \second{0.798} & 0.578 & \third{0.743} & 0.546 & 0.457 \\
    GPT-4o-mini & 0.722 & 0.554 & 0.507 & 0.608 & 0.321 \\
    Gemini-2.5-pro & 0.729 & \first{0.819} & 0.614 & \first{0.789} & \first{0.716} \\
    % Gemini-2.0-flash-think & ~ & ~ & ~ & ~ & ~ \\
    Gemini-2.0-flash & \first{0.830} & 0.638 & 0.643 & \third{0.696} & 0.543 \\
    Gemini-1.5-pro & 0.711 & 0.289 & 0.586 & 0.644 & 0.531 \\
    Gemini-1.5-flash & 0.671 & 0.494 & \second{0.779} & 0.536 & 0.444\\
    \midrule
    \textbf{Open-Source} & & & & \\
    GLM-4V-9B & 0.404 & 0.422 & 0.450 & \second{0.732} & 0.086 \\
    InternVL2-1B & 0.141 & 0.169 & 0.171 & 0.263 & 0.296 \\
    InternVL2-2B & 0.451 & 0.084 & 0.357 & 0.340 & 0.173 \\
    InternVL2-4B & 0.466 & 0.410 & 0.429 & 0.227 & 0.235 \\
    InternVL2-8B & 0.534 & 0.289 & 0.414 & 0.278 & 0.309\\
    InternVL2-26B & 0.592 & 0.108 & 0.543 & 0.510 & 0.037 \\
    InternVL2-40B & 0.567 & 0.494 & 0.429 & 0.309 & 0.222 \\ 
    InternVL2-76B & 0.549 & 0.386 & 0.257 & 0.407 & 0.309 \\ 
    InternVL2.5-1B & 0.329 & 0.253 & 0.379 & 0.381 &  0.309\\
    InternVL2.5-2B & 0.473 & 0.241 & 0.386 & 0.479 & 0.370 \\
    InternVL2.5-4B & 0.505 & 0.277 & 0.429 & 0.469 & 0.246 \\
    InternVL2.5-8B & 0.527 & 0.229 & 0.414 & 0.247 & 0.259\\
    InternVL2.5-26B & 0.635 & 0.277 & 0.571 & 0.418 & 0.346\\
    InternVL2.5-38B & 0.704 & 0.482 & 0.636 & 0.459 & 0.370\\
    InternVL2.5-78B & 0.507 & 0.578 & 0.507 & 0.474 & 0.407\\
    QwenVL-Chat & 0.458 & 0.277 & 0.214 & 0.356 & 0.309 \\
    Qwen2VL-2B & 0.343 & 0.277 & 0.214 & 0.562 & 0.506\\
    Qwen2VL-7B & 0.596 & 0.518 & 0.514 & 0.521 & 0.148 \\
    Qwen2VL-72B & 0.668 & 0.470 & 0.721 & 0.423 & 0.444 \\
    Qwen2.5-VL-3B & 0.567 & 0.325 & 0.329 & 0.521 & 0.222 \\
    Qwen2.5-VL-7B & 0.574 & 0.506 & 0.679 & 0.577 & 0.407 \\
    Qwen2.5-VL-32B & 0.635 & 0.506 & 0.457 & 0.500 & 0.494 \\
    Qwen2.5-VL-72B & 0.661 & 0.482 & 0.621 & 0.495 & \second{0.704} \\
    LLaVA-NeXT-7B & 0.466 & 0.108 & 0.271 & 0.505 & 0.247 \\
    Llama3.2-11B-VI & 0.585 & 0.410 & 0.486 & 0.665 & 0.210\\
    \midrule
    Random & 0.250 & 0.250 & 0.250 & 0.250 & 0.250 \\
    \bottomrule
    \end{tabular}
    }
    \caption{Performance comparison of various VLMs on our MCQs for the simulation tasks. The tasks include \textit{Place Carrot}, \textit{Close Drawer}, \textit{Straighten Rope}, \textit{Sweep Object}, and \textit{Ball Shooting}.}
    \label{tab:simulation_results}
\end{minipage}
\vspace{-8mm}
\end{table*}

\noindent\textbf{Results on MCQs from public robotic manipulation datasets.} Table~\ref{tab:bridge_droid_results} reports the performance of VLMs, random guessing, and human evaluation on the questions from public robotic manipulation datasets (Sec.~\ref{sec:q_gen_droid}). 
We use red to represent the highest accuracy, orange for the second‑highest, and yellow for the third‑highest, and we repeat this color code for subsequent tables. 
%The o1 and Gemini-2.5-pro models attain the best performance across all type 1 and type 2 questions for the Bridge dataset and the pick-and-place subset of the DROID dataset, indicating superior reasoning ability in pick-and-place tasks. 
The Gemini-2.5-pro model attains the best performance across all three type 1 question categories, while o1 performs the best at type 2 questions for DROID (p\&p). 
Among open-source VLMs, InternVL2.5-38B achieves the overall best performance across most questions. While the closed-source VLMs demonstrate a higher-than-random performance, all open-source VLMs with a size smaller than 2B perform similarly to random guessing, if not worse. See Appendix~\ref{appendix:addtional_results_public} for additional details.

% It also attains the best performance in the type 2 questions built on the DROID articulate subset. This may be due to its relatively large size and that it is the most recently-released model among the open-source VLMs we test.  

\begin{figure*}[t]
\begin{center}
\includegraphics[width=0.95\textwidth]{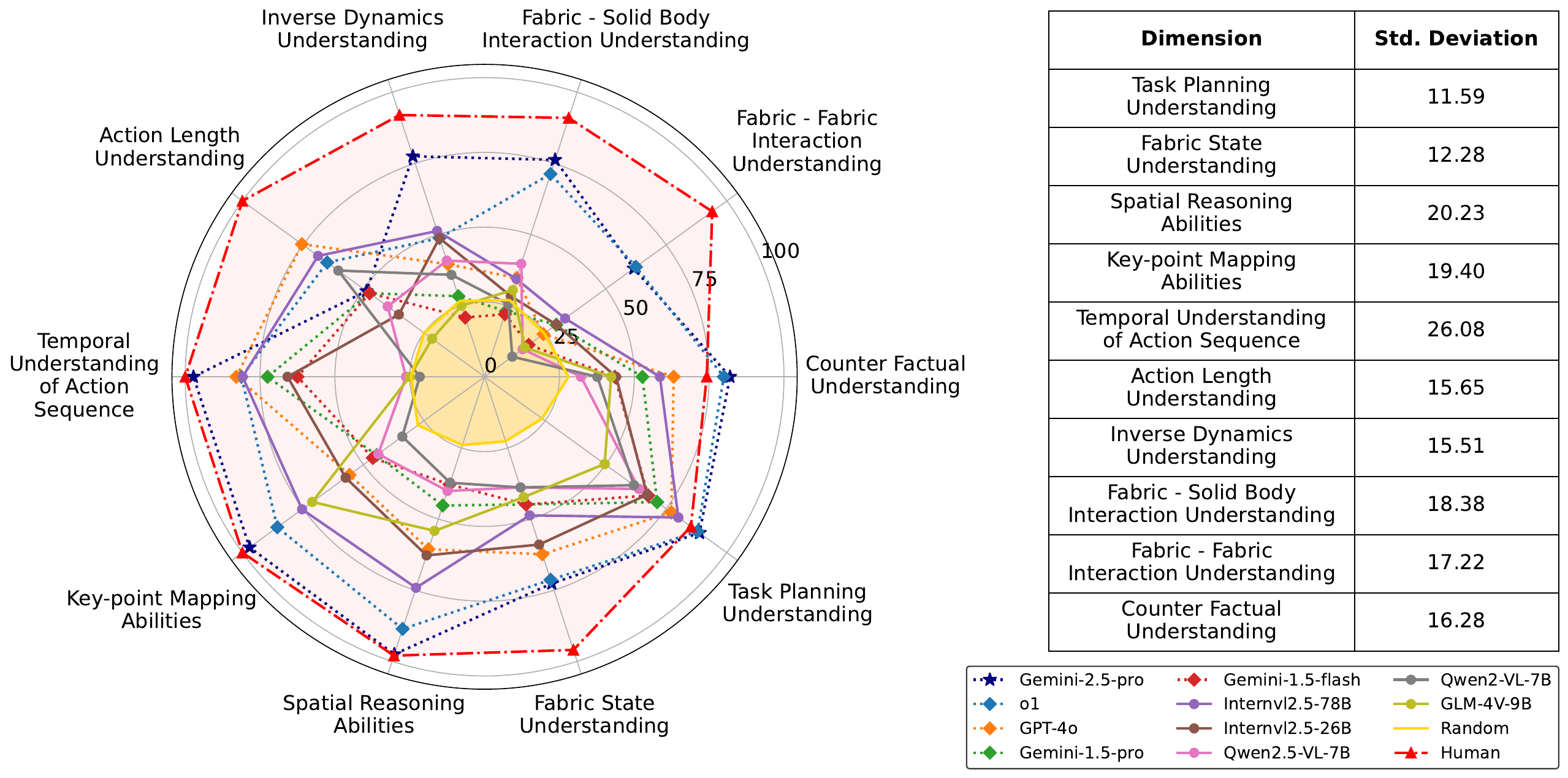}
\end{center}
\caption{
    The percentage accuracies of the VLMs for evaluating the dimensions of Fabric Manipulation, depicted as a Radar Chart, along with the standard deviation of their performances across these dimensions.
}
\label{fig:result_real_data_vedant}
\vspace{-1pt}
\end{figure*}

\noindent\textbf{Results on MCQs from in-house fabric manipulation setup.} See Fig.~\ref{fig:result_real_data_vedant} for a comparison of various VLMs' performance across multiple dimensions. Overall, VLMs outperform random guessing, suggesting a notable presence of physical reasoning capabilities in the context of fabric manipulation. Certain dimensions, like \textit{Task Planning Understanding}, are easier for all models, while others, like \textit{Fabric-Fabric Interaction Understanding}, are more challenging. However, consistent human accuracies across most dimensions suggest that the difficulty is not inherent, revealing gaps in VLMs' physical reasoning abilities. Among models, Gemini-2.5-pro consistently outperforms others, with o1 being a close contender.
We also observe that InternVL2.5-78B usually outperforms the smaller open-source models, though it trails behind Gemini-2.5-pro and o1. Furthermore, the high standard deviations of the model accuracies across different dimensions indicate their effectiveness in distinguishing the models' low-level inference abilities, particularly in dimensions like \textit{Temporal Understanding of Action Sequence} and \textit{Spatial Reasoning Abilities}, justifying our choice of dimensions. Performance of all models on these questions is detailed in  Appendix~\ref{appendix:additional_results_fabric}.

\noindent\textbf{Results on MCQs from simulation.} Table~\ref{tab:simulation_results} reports the performance of VLMs on MCQs from simulation. While Gemini-2.5-pro still has the overall best performance among closed-source models, we observe that certain models excel in specific tasks. For instance, Gemini-2.0-flash achieves the best performance on the \textit{Place Carrot} task, suggesting that some models may be better optimized for certain tasks. Among open-source models, larger variants frequently outperform smaller ones in the same family, though the rate of improvement varies based on the task. 
%For example, in \textit{Straight Rope}, the InternVL-2.5 series stops following the near-linear improvement one would expect, signaling a plateau in its scaling behavior.  

% We observed scenarios where smaller models, exhibiting relatively lower performance in most tasks, outperform the otherwise stronger models in particular tasks, suggesting that some models are better optimized for some tasks (for instance, Gemini-1.5-flash outperforms GPT-4o in \emph{Straighten Rope} task).

% \noindent\textbf{Comparison between model performance across data sources.} We notice that the model's relative standing can reverse between data sources. For instance, the pattern of a given model's performance across different data sources do not perfectly align. 

% Daniel: note to self, only uses questions with random guessing 0.25 here.
\noindent\textbf{Manipulation task category breakdown.} \benchmark covers five task categories: pick-and-place, articulated object, deformable object, tool, and dynamic manipulation. Aggregating over all model accuracies in each task category (w/o Q2 MCQs from public datasets) reveals that existing VLMs handle pick-and-place comparatively well, achieving a mean accuracy of $0.525$. %(for Q1 MCQs). 
In contrast, articulated object and dynamic manipulation are more challenging, with accuracies of $0.396$ and $0.357$.

\noindent\textbf{Model robustness across task categories.} We compute each model's coefficient of variation (CV), across all manipulation task categories to analyze its robustness, an ability to perform accurately in different contexts (lower CV values are better). We find that Gemini-2.5-pro is the most robust closed-source model with a CV of $0.089$. Qwen2.5-VL-32B is the most robust open-source model ($0.085$); while it has slightly worse average accuracy than its larger variant (Qwen2.5-VL-72B with a CV of $0.133$), it has smaller standard deviation. 
% We also observe that InternVL2.5-38B, while not uniformly strong on all tasks, excels at pick-and-place tasks. % For the cost-conscious teams we suggest Qwen2.5-VL-7B, that has uniformly decent performance across all tasks, with a average accuracy of $0.517$ and SD of $0.1147544279$.

% \subsection{Fabric Manipulation Specific Factorized Dimension Analysis}

% \subsection{Scaling Law for Open-source Models}

% \subsection{Error Modes}

% Daniel: I think the section title should have Real World somewhere.
\subsection{Performance Transfer to Unseen Real World Robot Experiments}
\label{ssec:real_world_results}
% Among the closed-source models, o1 and GPT-4o achieve the highest success rates while Gemini-1.5-pro significantly under-performs, showing similar accuracy to InternVL2.5-26B (see Table~\ref{tab:Real-world-results}). Smaller models like InternVL2.5-8B outperform GLM-4V but still lag behind the closed-source models. In the \emph{Move Fabric Hard} task, where VLMs must reason about fabric movement without causing wrinkles or folds, all models perform poorly. In addition, identical success rates across VLMs do not imply identical failure modes. For instance, Gemini-1.5-pro and InternVL2.5-26B both achieve 1/3 success in the \emph{Pick Crumpled Cloth} task but fail for different reasons.

To demonstrate the utility of \benchmark, we evaluate a set of selected VLMs on 7 unseen real world manipulation tasks (Sec.~\ref{sec:evaluation}). We perform statistical analysis to assess the correlation between VLM performance on our MCQ benchmark and their effectiveness as real-world robotic agents (Table.~\ref{tab:Real-world-results}). On aggregating the accuracies over the different question-types, we observe a Pearson’s coefficient of 0.889 ($p = 0.003$), a Spearman’s coefficient of 0.850 ($p = 0.007$), and a Kendall’s Tau of 0.691 ($p = 0.018$). These results indicate a strong positive correlation between MCQ benchmark performance and real-world effectiveness. The statistically significant ($p < 0.05$) Pearson’s coefficient highlights strong linear alignment, while the Spearman and Kendall coefficients confirm robust monotonic agreement with high confidence in rank-order consistency. Collectively, these trends support the utility of \benchmark as a reliable proxy for evaluating VLMs in embodied robotic settings. See Appendix~\ref{appendix:additional_results_stats} for details on correlations for each question type with the real-world experiments.

% We use Pearson’s and Spearman’s coefficients to capture linear and monotonic relationships, respectively. Each coefficient is accompanied by a p-value, with values below 0.05 usually indicating statistical significance.  

% \subsection{Key Insights}
% % \noindent\textbf{For practitioners.} Based on our extensive evaluation and analysis, we 
% % \hejia{Which model is the "safest" option}
% % \hejia{if I only want to work on one task modality, do I need a giant closed-source model?}
% % \hejia{Do bigger models always pay off}
% % \noindent\textbf{For VLM researchers.} 
% % \hejia{which task is the bottleneck?}
% % \hejia{is scaling plateauing?}
% % \hejia{is spatial reasoning still weak? what error modes tell us? picking is easy, placing is harder?}
% % \hejia{robustness or peak accuracy? which matters more?}

% Our extensive evaluation indicates three take-aways: (i) Gemini-2.5-Pro or GPT-o1 are the reliable ``set-and-forget'' options for general manipulation reasoning. (ii) cost-conscious teams concerned only with prehensile pick-and-place manipulation can rely on InternVL-4B, which delivers near-top accuracy at a fraction of the price.  (iii) Deformable object manipulation remains the dominant, signaling a prime target for further research.   

\begin{table*}[t]
\scriptsize
\begin{center}
\resizebox{\textwidth}{!}{%
\begin{tabular}{lccccccc|c}

\toprule
 & \multicolumn{1}{c}{\bf Color} & \multicolumn{1}{c}{\bf Spatial} & \multicolumn{1}{c}{\bf Fit} & \multicolumn{1}{c}{\bf Pick} & \multicolumn{1}{c}{\bf Move} & \multicolumn{1}{c}{\bf Move} & \multicolumn{1}{c}{\bf Move} & \multicolumn{1}{c}{\bf Total} \\

 \multicolumn{1}{l}{\bf Model}& \multicolumn{1}{c}{\bf Specific} & \multicolumn{1}{c}{\bf Specific}& \multicolumn{1}{c}{\bf Apple} & \multicolumn{1}{c}{\bf Crumpled} & \multicolumn{1}{c}{\bf  Fabric} & \multicolumn{1}{c}{\bf  Fabric} & \multicolumn{1}{c}{\bf  Fabric} & \multicolumn{1}{c}{\bf Success} \\

  & \multicolumn{1}{c}{\bf p\&p} & \multicolumn{1}{c}{\bf p\&p} &   & \multicolumn{1}{c}{\bf Cloth} &\multicolumn{1}{c}{\bf Easy}  & \multicolumn{1}{c}{\bf Intermediate} & \multicolumn{1}{c}{\bf Hard}& \multicolumn{1}{c}{\bf Rate} \\

\midrule
o1                     & 1/3 & 3/3 & 2/3 & 3/3 & 3/3 & 3/3 & 0/3 & \second{15/21} \\
GPT-4o                 & 2/3 & 2/3 & 3/3 & 3/3 & 3/3 & 0/3 & 1/3 & \third{14/21} \\
Gemini-2.5-pro         & 3/3 & 3/3 & 3/3 & 3/3 & 3/3 & 3/3 & 0/3 & \first{18/21} \\
Gemini-1.5-pro         & 2/3 & 0/3 & 0/3 & 1/3 & 2/3 & 0/3 & 0/3 & 5/21  \\
InternVL2.5-78B        & 2/3 & 2/3 & 1/3 & 2/3 & 3/3 & 2/3 & 0/3 & 12/21 \\
InternVL2.5-26B        & 1/3 & 2/3 & 2/3 & 1/3 & 2/3 & 3/3 & 0/3 & 8/21  \\
InternVL2.5-8B         & 2/3 & 2/3 & 0/3 & 0/3 & 1/3 & 0/3 & 0/3 & 5/21  \\
GLM-4V-9B              & 1/3 & 0/3 & 0/3 & 1/3 & 0/3 & 0/3 & 0/3 & 2/21  \\

\bottomrule
\end{tabular}
}
\caption{
Performance comparisons for the real-world experiments with different VLMs via the success rate. 
% (``Total Success Rate''). and the accuracy of the VLMs' predictions (``Total Correct Prediction''). 
% The latter is higher than the former, as some task failures stem from hardware limitations instead of VLMs' decision-making.
}
\label{tab:Real-world-results}
\end{center}
\vspace{-10pt}
\end{table*}

%===============================================================================

\section{Conclusion}
\label{sec:conclusion}

In this work, we propose a novel benchmark, \benchmark. This is a robotics-focused benchmark which critically analyzes modern VLMs and their ability to reason about object properties and precise movements for robotic manipulation.
Despite the presence of better than random reasoning capabilities, our results indicate that robot manipulation understanding across various models is relatively poor, leaving much room for improvement. 
We plan to maintain this benchmark and to further improve it by addressing some of its current limitations. We hope that this work helps to facilitate a better understanding of VLMs as they continue to play a bigger role in robotic manipulation. 

%===============================================================================

% Daniel: moving this here since limitations are now after the 8-page limit.
\clearpage

\section{Limitations}
\label{sec:limitations}

While we believe \benchmark is a valuable benchmark for VLMs and robotics, there are some limitations that suggest opportunities for future work. 
First, due to limited computational and financial resources, we did not conduct experiments on all possible closed-source and open-source model variants, which may draw an incomplete conclusion. %when comparing against state-of-the-art closed-source models \daniel{hopefully this changes}. 
Second, MCQ evaluations depend on access to potential correct/incorrect options to choose from. In real-world scenarios where VLMs are employed as agents, they might not always be able to choose among pre-selected options. 
Third, the benchmark is not exhaustive across different tasks and excludes crucial deformable manipulation tasks that can require low-level reasoning, such as robotic bag manipulation~\cite{autobag2023,shi2024yell,ALOHA}. Fourth, although we demonstrate a strong correlation between benchmark performance and real-world experiment performance, our current setup leverages VLMs in the same way across both settings where the VLMs select an answer or action from a predefined list of options. A natural next step would have been to explore alternative approaches for real-world experiments, such as requiring VLMs to generate actions directly rather than choosing from fixed options. Finally, \benchmark lacks emphasis on higher-level aspects (e.g., task planning) since it is focused on low-level reasoning.

%===============================================================================

% The acknowledgments are automatically included only in the final and preprint versions of the paper.
\acknowledgments{
We thank our colleagues who gave us helpful feedback, including Minjune Hwang and Rajas Chitale. We used LLMs to help with proofreading our writing and manually verified all such content. 
}

%===============================================================================

\bibliography{references}  % .bib

%===============================================================================

\clearpage
% \end{document}
\appendix

\addcontentsline{toc}{section}{Appendix} % Add the appendix text to the document TOC
\part{Appendix} % Start the appendix part
\mtcsetdepth{parttoc}{3} % show subsubsections
\parttoc % Insert the appendix TOC
\newpage

\section{Additional Details about Table~\ref{tab:pull}}
\label{appendix:table}

In Table~\ref{tab:pull}, we compare \benchmark with 9 leading vision-language benchmarks or datasets~\cite{wang2023newton,PhysBench2025,Octopi2024,zhang2024vlabenchlargescalebenchmarklanguageconditioned,gao2024physicalVLM,guruprasad2024benchmarkingvisionlanguage,10802733,fu2024can,garcia24gembench} in the following 7 dimensions: 
\begin{enumerate}[leftmargin=*,noitemsep]
\item \textit{Low-level Manipulation Reasoning}, assessing whether included evaluations demand precise action-centric reasoning rather than solely high-level judgments (e.g., object attributes, task planning, or spatial relationships).
\item \textit{Manipulation Task Factorization}, indicating whether evaluations are decomposed along orthogonal axes (e.g., required manipulation skills, fabric state understanding, spatial reasoning) to enable systematic analysis.
\item \textit{Performance Transfer Evaluation}, denoting the presence of dedicated splits for measuring generalization to unseen task variations.
\item \textit{Real World Data}, referring to the inclusion of evaluation questions sourced from real-world robot manipulation setups.
\item \textit{Fabric Manipulation}, referring to the inclusion of evaluations on fabric manipulation tasks.
\item \textit{Dynamic Manipulation}, referring to the inclusion of evaluations on dynamic manipulation tasks.
\item \textit{Model Diversity}, measuring the breadth of models evaluated: “low” if fewer than five, “high” if more than ten.
\end{enumerate}

For datasets such as PhysObjects~\cite{gao2024physicalVLM}, Octopi~\cite{Octopi2024}, and GemBench~\cite{garcia24gembench}, several fields remain blank because these resources were introduced as data collections rather than comprehensive benchmarks.

\section{Additional Details about Data Preparation}
\label{appendix:data_preparation}

\subsection{From Public Robotic Manipulation Datasets}
\label{appendix:data_preparation_public}

To effectively utilize data from the existing public robotic manipulation datasets of DROID~\cite{khazatsky2024droid} and Bridge~\cite{walke2023bridgedata} for \benchmark, we require accurate gripper positions to obtain the picking and placing points. Although DROID provides RGBD images and the camera intrinsic matrix for calculating the gripper's position, we found a notable empirical difference from the calculation to the ground truth.
Hence, we utilize the annotated DROID subset from~\cite{kuang2024ramretrievalbasedaffordancetransfer}, retaining only the successful rollouts for question generation. Additionally, we employ the Bridge-V2~\cite{walke2023bridgedata} dataset from the Open-X collection~\cite{open_x_embodiment_rt_x_2023}, sampling 450 successful rollouts. We exclude 96 rollouts in total, of which 45 is due to how the Grounding DINO model fails to detect the key object $O_K$. In the other 51 rollouts, either the task description does not match the rollout video or it is not a pick-and-place task.

\begin{figure*}[t]
\begin{center}
\includegraphics[width=.98\textwidth]{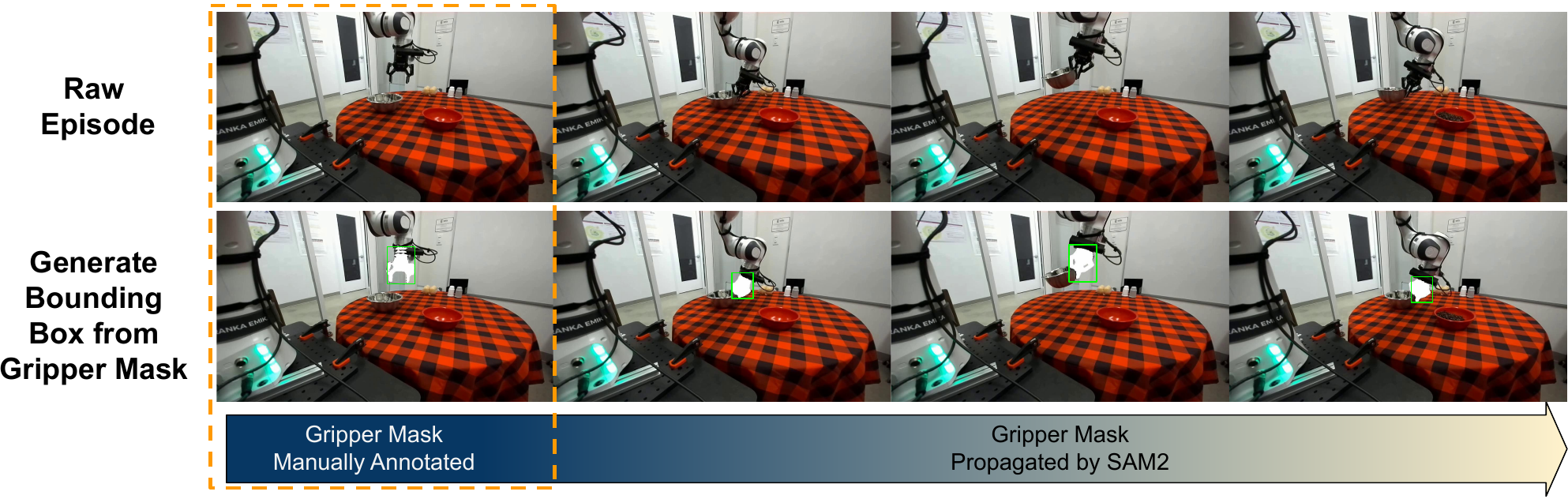}
\end{center}
\caption{Raw frames (top) are paired with gripper masks (bottom), initially annotated manually (see orange box) and then propagated using SAM2~\cite{ravi2024sam2}. Bounding boxes derived from the masks are used for fine-tuning Grounding DINO~\cite{liu2024grounding}.}
\label{fig:finetune_GD}
\end{figure*}

We further separate the DROID data into two subsets based on the manipulation task. In the ``articulate manipulation'' subset, the gripper trajectories are complete and thus we directly use them. However, in the ``pick-and-place manipulation'' subset, only the gripper trajectory in the picking phase is provided. 
To complete the trajectories from~\citep{kuang2024ramretrievalbasedaffordancetransfer}, we fine-tune Grounding DINO~\citep{liu2024grounding} to detect the gripper, as shown in Fig.~\ref{fig:finetune_GD}.
We annotate about 50 episodes (each with 2 to 10 frames) to fine-tune the Grounding DINO model. For each episode, we manually provide the gripper mask in one frame and then use Segment Anything 2 (SAM2)~\citep{ravi2024sam2} to track the gripper throughout the episode to efficiently generate image-mask pairs. We extract the gripper trajectories in the Bridge data in a similar fashion.

\subsection{From In-house Fabric Manipulation Setup}
\label{appendix:data_preparation_fabric}

As described in Table~\ref{tab:fabric_dimensions}, we break down common aspects of fabric manipulation into ten distinct dimensions. This list is not exhaustive and does not encompass the entire range of fabric manipulation tasks~\cite{longhini2024unfoldingliteraturereviewrobotic}. However, we believe a strong understanding of these dimensions correlates with fabric manipulation performance.

For instance, understanding how different objects in a scene interact with each other is crucial when performing fabric manipulation tasks in cluttered environments. Furthermore, the nature of this interaction also varies significantly between rigid objects and deformable objects. To evaluate this, we collect image observations of multiple scenes with certain objects randomly placed in the scene. Given an image, the robot agent should reason about whether or not it is possible to achieve the desired fabric configuration (described via natural language) and if so, then how. 

Additionally, understanding inverse dynamics is important when reasoning about feasible pick-and-place action(s) to perform goal-based tasks like fabric folding. For this dimension, we collect image observations of a fabric on a table and record the initial and final fabric configurations after performing a robot action like folding or unfolding. An agent should be able to generate the correct robot action from these images across diverse settings.

Similarly, it is essential for a robot agent to reason about the configuration of the fabric from its observation and about the consequences of an action if the scene were to slightly change. Moreover, it might not be trivial for a robot agent to demonstrate spatial reasoning capabilities, a foundation for numerous tasks~\cite{PhysBench2025}. For evaluating VLMs along these dimensions, we collect a single image observation per scene and manually label them with information required for further question generation. For instance, we might label the locations of different corners of a flat fabric on a table.

% Daniel (Jan 31): removing the fleets here since we don't study that. Also, added the emphasis on low-level reasoning here, and did some general trimming. 
While our main focus is on emphasizing low-level reasoning capabilities, we also study some high-level reasoning. This refers to the ability to decompose complex tasks into smaller and/or more manageable sub-tasks. To this end, we consider certain aspects of high-level task planning and maintain a list of common sense statements (e.g., folding a fabric will be more effective if it is flattened first) which are used later to generate multiple-choice questions.

\begin{table}[t]
    \centering
    \renewcommand{\arraystretch}{1.3} % Adjust row height
    \setlength{\tabcolsep}{6pt} % Adjust column spacing
    \small % Reduce font size for compactness
    \resizebox{\linewidth}{!}{ % Ensure table fits within margins
    \begin{tabularx}{\linewidth}{|l|X|}
        \hline
        \textbf{Dimension} & \textbf{Description} \\
        \hline
        \textit{Task Planning Understanding} & 
        Assesses physical reasoning based on human intuition, requiring VLMs to select the correct answer. \\
        \hline
        \textit{Fabric State Understanding} & 
        Tests VLMs' ability to identify the correct fabric state from an image and four given options. \\
        \hline
        \textit{Spatial Reasoning Abilities} & 
        Evaluates VLMs' ability to locate fabric corners (e.g., “bottom-right”) from an image. \\
        \hline
        \textit{Key-point Mapping Abilities} & 
        Assesses VLMs' accuracy in mapping key points from an image to a grid location. \\
        \hline
        \textit{Temporal Understanding of Action Sequence} & 
        Tests VLMs' ability to reorder shuffled images of a fabric manipulation sequence correctly. \\
        \hline
        \textit{Action Length Understanding} & 
        Evaluates VLMs' understanding of how short vs. long pick-place actions affect fabric configuration. \\
        \hline
        \textit{Inverse Dynamics Understanding} & 
        Requires VLMs to predict the correct pick-place action from four choices based on initial and final images. \\
        \hline
        \textit{Fabric-Solid Body Interaction Understanding} & 
        Tests VLMs' ability to choose the correct pick-place action when fabric interacts with solid objects. \\
        \hline
        \textit{Fabric-Fabric Interaction Understanding} & 
        Assesses VLMs' understanding of multi-fabric interactions, especially in bi-manual actions. \\
        \hline
        \textit{Counterfactual Understanding} & 
        Evaluates VLMs' reasoning on how changes in a scene or action alter outcomes. \\
        \hline
    \end{tabularx}
    }
    \vspace{5pt}
    \caption{Description of the different dimensions for fabric manipulation, evaluated as a part of \benchmark. See Table~\ref{tab:num-questions} for the number of corresponding questions in \benchmark.}
    \label{tab:fabric_dimensions}
    \vspace{-10pt}
\end{table}

\subsection{From Existing Simulation Environments}
\label{appendix:data_preparation_sim}

\begin{figure*}[t]
\begin{center}
\includegraphics[width=.98\textwidth]{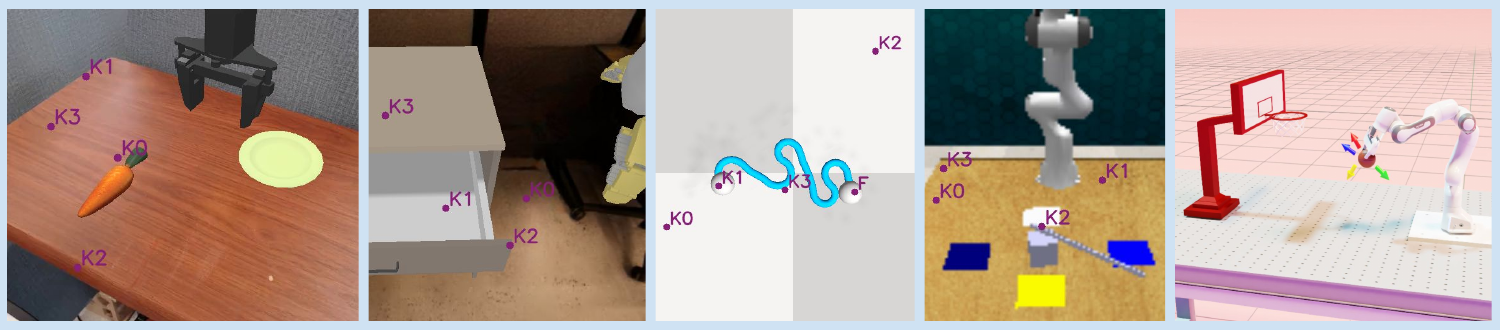}
\end{center}
\caption{
Simulated tasks from ManipBench: \textit{Place Carrot}, \textit{Close Drawer}, \textit{Straighten Rope}, \textit{Sweep Object}, \textit{Ball Shooting}. For the first four tasks, we use the keypoints (\texttt{K0, K1, K2, K3}) generated from demonstrations to encode robot actions and enable multi-choice question design. For the last task, we use colored arrows (red, green, yellow, blue) to encode robot actions.
}
\label{fig:sim_tasks}
\vspace{-5pt}
\end{figure*}

As discussed in Section~\ref{section:data}, for each task category, we either reuse simulation environments from prior benchmarks (Fig.~\ref{fig:sim_tasks}, first four images) or develop custom simulation environments (Fig.~\ref{fig:sim_tasks}, last image). Demonstrations are then generated using rollouts of pre-trained policies or via manual annotations, following a task-specific data collection procedure.

\begin{enumerate}[leftmargin=*,noitemsep]
    \item \textit{Place Carrot (pick-and-place).} We use the simulation environment and pre-trained Octo policy~\cite{octo_2024} and RT-1 policy~\cite{rt12022arxiv} from SimplerEnv to generate a set of robot demonstrations. We then identify keypoints representing the ground-truth pick-and-place actions using contact information provided by the simulator. 
    \item \textit{Close Drawer (articulated object manipulation).} Similar to the Place Carrot task, we use SimplerEnv to generate robot demonstrations. For this task, we identify keypoints that represent the robot-drawer contact point and the robot movement direction following contact.
    \item \textit{Straighten Rope (deformable object manipulation).} We use SoftGym to simulate rope dynamics and generate demonstrations using a heuristic: the robot grippers pull two endpoints of the rope apart. In this task, keypoints represent the robot-rope contact point, as well as the robot movement direction following contact.
    \item \textit{Sweep Object (tool manipulation).} We use the environment from RLBench and pre-trained policy from PerAct~\cite{shridhar2022peract} to generate robot demonstrations. The keypoints represent the tool-robot contact point, tool-object contact point, and the robot movement direction following contact.
    \item \textit{Ball Shooting (dynamic manipulation).} We create our own ball-shooting simulation environments in IsaacSim~\cite{isaacsim}. We manually annotate colored arrows on the images to denote different ball-shooting directions for the robot. 
\end{enumerate}

\section{Additional Details about Question Generation}
\label{appendix:question_generation}

Using the different data sources as described in Sec.~\ref{section:data}, we generate \textbf{\QuestionNum} multiple-choice questions. See Table~\ref{tab:num-questions} for the question statistics.

\subsection{From Public Robotic Manipulation Datasets}
\label{appendix:question_generation_public}

We generate questions across 612 tasks as described in Section~\ref{sec:q_gen_droid}. Our preprocessing phase will sample 18 incorrect points for each episode and we only need 3 for building a question (with the 4th point being the ``correct'' choice). Thus, to further exploit the dataset we have, we perform ``question augmentation'' by randomly sampling different sets of 3 incorrect points to build several question versions from the same episode. We also randomly sample 3 pairs of starting and ending tiles that are distinct from the ground-truth tiles as the incorrect options. 

\begin{table*}[t]
\centering
\begin{minipage}[t]{0.53\linewidth}
    \centering
    \footnotesize
    \begin{tabular}{@{}p{5.7cm} >{\raggedleft\arraybackslash}p{1.35cm}@{}}
    \toprule
    \textbf{Task Type} & \textbf{Total Questions} \\ 
    \midrule
    \multicolumn{2}{@{}l}{\textbf{From Public Robotic Datasets}} \\ 
    \hspace{0cm}\textit{Question Type 1} & \\ 
    \hspace{0.5cm}DROID pick and place (p\&p) & 2010 \\ 
    \hspace{0.5cm}DROID articulated (art.) & 1640 \\ 
    \hspace{0.5cm}Bridge & 2470 \\ 
    \hspace{0cm}\textit{Question Type 2} & \\ 
    \hspace{0.5cm}DROID pick and place (p\&p) & 1005 \\ 
    \hspace{0.5cm}DROID articulated (art.) & 820 \\ 
    \hspace{0.5cm}Bridge pick and place & 1235 \\ 
    \midrule
    \multicolumn{2}{@{}l}{\textbf{For Evaluating Fabric Manipulation}} \\ 
    \hspace{0cm}Task Planning Understanding & 240 \\ 
    \hspace{0cm}Fabric State Understanding & 234 \\ 
    \hspace{0cm}Spatial Reasoning Abilities & 325 \\ 
    \hspace{0cm}Keypoint Mapping Abilities & 312 \\ 
    \hspace{0cm}Temporal Understanding of Action Sequence & 240 \\ 
    \hspace{0cm}Action Length Understanding & 240 \\ 
    \hspace{0cm}Inverse Dynamics Understanding & 240 \\ 
    \hspace{0cm}Fabric-Solid Body Interaction Understanding & 282 \\ 
    \hspace{0cm}Fabric-Fabric Interaction Understanding & 280 \\ 
    \hspace{0cm}Counterfactual Understanding & 269 \\ 
    \midrule
    \multicolumn{2}{@{}l}{\textbf{From Existing Simulation Environments}} \\ 
    \hspace{0cm}Place Carrot (pick-and-place task) & 277 \\ 
    \hspace{0cm}Close Drawer (articulated manipulation task) & 83 \\ 
    \hspace{0cm}Straighten Rope (deformable manipulation) & 140 \\ 
    \hspace{0cm}Sweep Object (tool manipulation task) & 194 \\ 
    \hspace{0cm}Ball Shoot (dynamic manipulation task) & 81\\ 
    \bottomrule
    \textbf{All Tasks Combined} & \QuestionNum \rule[-0.5mm]{0pt}{4mm} \\
    \bottomrule
    \end{tabular}
    \caption{
        Summary of tasks and the number of questions in each category. \benchmark has \textbf{\QuestionNum} questions in all. See Sec.~\ref{section:data} and Sec.~\ref{sec:gen_questions} for more details about the data sources and the question generation process, respectively.
    }
    \label{tab:num-questions}
\end{minipage}%
\hfill
\begin{minipage}[t]{0.45\linewidth}
    \centering
    \footnotesize
    \begin{tabular}{@{}p{4.4cm}p{1.0cm}<{\raggedleft}@{}}
    \toprule
    \textbf{Model Family} & \textbf{Params} \\ 
    \midrule
    \multicolumn{2}{@{}l}{\textbf{OpenAI GPT~\cite{GPT-4} (closed-source)}} \\ 
    \hspace{0.5cm}o1 & N/A \\
    \hspace{0.5cm}GPT-4.1 & N/A \\
    \hspace{0.5cm}GPT-4o & N/A \\ 
    \hspace{0.5cm}GPT-4o-mini & N/A \\ 
    \midrule
    \multicolumn{2}{@{}l}{\textbf{Google Gemini~\cite{Gemini} (closed-source)}} \\ 
    \hspace{0.5cm}Gemini-2.5-pro & N/A \\ 
    % \hspace{0.5cm}Gemini-2.0-flash-think & N/A \\ 
    \hspace{0.5cm}Gemini-2.0-flash & N/A \\ 
    \hspace{0.5cm}Gemini-1.5-pro & N/A \\ 
    \hspace{0.5cm}Gemini-1.5-flash & N/A \\ 
    \midrule
    \multicolumn{2}{@{}l}{\textbf{GLM-4V~\cite{wang2023cogvlm,glm2024chatglm} (open-source)}} \\ 
    \hspace{0.5cm}GLM-4V-9B & 13.9B \\ 
    \midrule
    \multicolumn{2}{@{}l}{\textbf{InternVL-2~\cite{chen2024far, chen2023internvl} (open-source)}} \\ 
    \hspace{0.5cm}InternVL-2-1B & 0.94B \\ 
    \hspace{0.5cm}InternVL-2-2B & 2.21B \\ 
    \hspace{0.5cm}InternVL-2-4B & 4.15B \\ 
    \hspace{0.5cm}InternVL-2-8B & 8.08B \\ 
    \hspace{0.5cm}InternVL-2-26B & 25.50B \\ 
    \hspace{0.5cm}InternVL-2-40B & 40.10B \\ 
    \hspace{0.5cm}InternVL-2-76B & 76.30B \\ 
    \midrule
    \multicolumn{2}{@{}l}{\textbf{InternVL-2.5~\cite{internvl-2.5} (open-source)}} \\ 
    \hspace{0.5cm}InternVL-2.5-1B & 0.94B \\ 
    \hspace{0.5cm}InternVL-2.5-2B & 2.21B \\ 
    \hspace{0.5cm}InternVL-2.5-4B & 3.71B \\ 
    \hspace{0.5cm}InternVL-2.5-8B & 8.08B \\ 
    \hspace{0.5cm}InternVL-2.5-26B & 25.50B \\
    \hspace{0.5cm}InternVL-2.5-38B & 38.40B \\ 
    \hspace{0.5cm}InternVL-2.5-78B & 78.40B \\
    \midrule
    \multicolumn{2}{@{}l}{\textbf{Qwen-VL~\cite{bai2023qwen} (open-source)}} \\ 
    \hspace{0.5cm}Qwen-VL-Chat-Int4 & 4.05B \\ 
    \hspace{0.5cm}Qwen-VL-Chat & 9.60B \\ 
    \midrule
    \multicolumn{2}{@{}l}{\textbf{Qwen2-VL~\cite{wang2024qwen2} (open-source)}} \\ 
    \hspace{0.5cm}Qwen2-VL-2B   & 2.21B \\ 
    \hspace{0.5cm}Qwen2-VL-7B   & 8.29B \\
    \hspace{0.5cm}Qwen2-VL-72B  & 73.40B \\
    \midrule
    \multicolumn{2}{@{}l}{\textbf{Qwen2.5-VL~\cite{bai2025qwen2} (open-source)}} \\
    \hspace{0.5cm}Qwen2.5-VL-3B & 3.75B \\
    \hspace{0.5cm}Qwen2.5-VL-7B & 8.29B \\
    \hspace{0.5cm}Qwen2.5-VL-32B & 33.50B \\
    \hspace{0.5cm}Qwen2.5-VL-72B & 73.40B \\    
    \midrule
    \multicolumn{2}{@{}l}{\textbf{LLaVA-NeXT~\cite{liu2023llava, liu2023improvedllava, liu2024llavanext} (open-source)}} \\ 
    \hspace{0.5cm}LLaVA-NeXT-7B & 7.57B \\ 
    \midrule
    \multicolumn{2}{@{}l}{\textbf{Llama3.2-Vision~\cite{touvron2023llama,grattafiori2024llama3herdmodels} (open-source)}} \\ 
    \hspace{0.5cm}Llama3.2-11B-Vision-Instruct & 10.60B \\
    \bottomrule
    \end{tabular}
    \caption{
        Summary of model families evaluated.
        %with open and closed-source models. See Sec.~\ref{sec:evaluation}.
    }
    \label{tab:num-models}
\end{minipage}
\vspace{-5pt}
\end{table*}

\subsection{From In-house Fabric Manipulation Setup}
\label{appendix:question_generation_fabric}

We describe how we generate the MCQs for evaluating fabric manipulation in Sec.~\ref{sec:q_gen_fabric}. On having all the data mapped in terms of affordances, we formulate different types of multiple-choice questions (see Table~\ref{tab:fabric_dimensions}). We use the collected image observations to formulate the questions and have the obtained affordances act as the correct answers. The incorrect options for these questions are generated manually, making sure that they are clearly distinct from the answer. For instance, for the question types whose choices are based on the possible grid cells, we make sure that the cells neighboring the cell corresponding to the correct answer are not included in the possible choices.

\subsection{From Existing Simulation Environments}
\label{appendix:question_generation_sim}

\begin{figure}[t]
\center
\includegraphics[width=\textwidth]{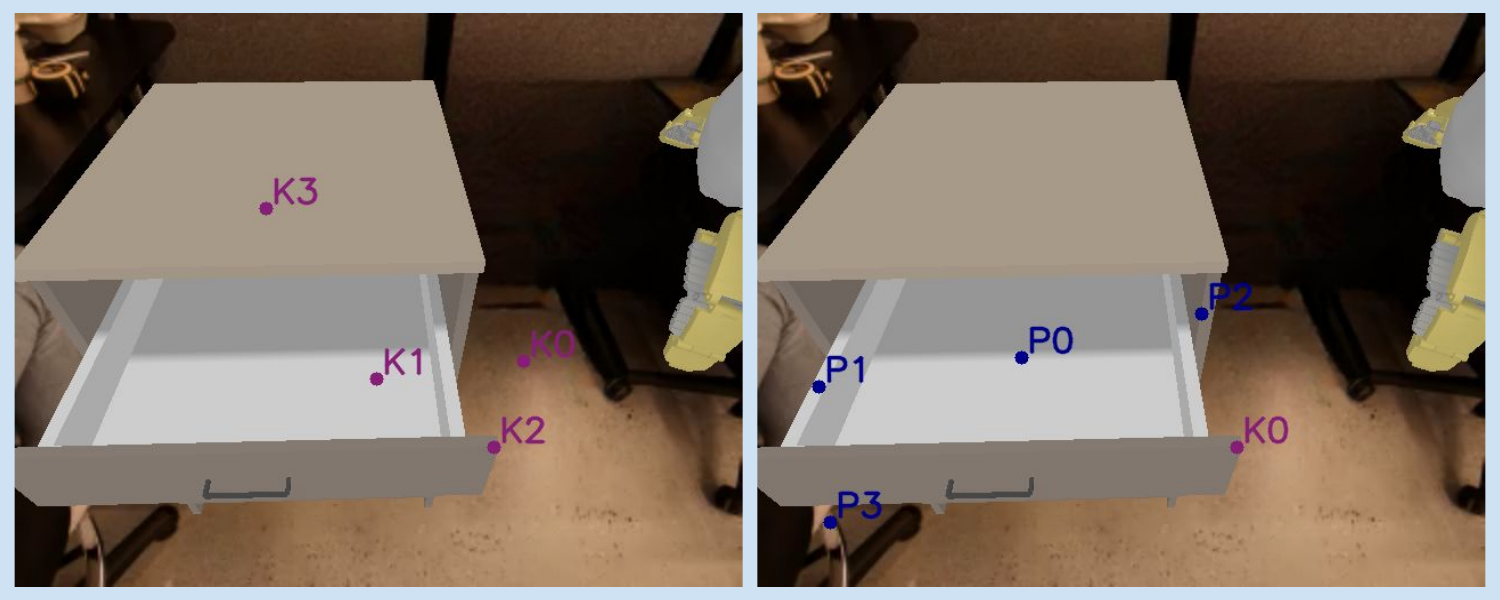}
\caption{
    Two annotated images illustrating evaluation question generation for VLMs in the \textit{Close Drawer} task. Left: Keypoints denote potential contact points for initiating the drawer-closing action. Right: Keypoints indicate possible movement directions for closing the drawer while maintaining contact at point $K_0$. 
}
\label{fig:question_gen_sim}
\vspace*{-10pt}
\end{figure}

After generating robot demonstrations from simulation (Sec.~\ref{section:data}), we create evaluation questions for VLMs. In each demonstration, we identify the frames $f_s$ and $f_e$ where manipulation ``starts" and ``ends," respectively. We then annotate the candidate contact points and after-contact movements on the frames to generate evaluation questions. VLMs are required to select from the candidate keypoints to complete the given manipulation tasks. For instance, in Fig.~\ref{fig:question_gen_sim}, they must choose one contact point from \texttt{K0} to \texttt{K3} to initiate contact with the drawer and one point from \texttt{P0} to \texttt{P3} to push the drawer. 

In our ball-shooting task (Fig.~\ref{fig:sim_tasks}, right), instead of keypoint-based action annotations, we use colored arrows to encode robot shooting actions. VLMs must choose one arrow as the robot shooting direction. We designed three types of questions to evaluate the dynamic manipulation reasoning capabilities of VLMs in the ball-shooting task (see Sec.~\ref{sec:ball_shoot}).

\section{Additional Details about Human Evaluation}
\label{appendix:human_evaluation}

\begin{figure}[t]
\center
\includegraphics[width=0.8\textwidth]{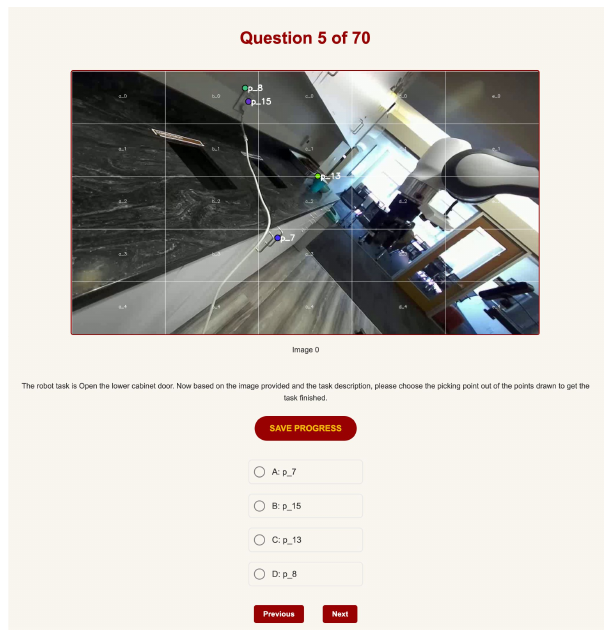}
\caption{
Visualization of the website used for human evaluation. The human selects one of the given choices.
}
\label{fig:human_eval}
\vspace*{-10pt}
\end{figure}

As described in Sec.~\ref{sec:evaluation}, we perform a web-based human evaluation of our MCQs. We design the website using Python flask and host it using \texttt{ngrok} operated servers. We make the website easy for the volunteers to navigate through the questions, along with incorporating a feature that allows them to save their responses for later (see Fig.~\ref{fig:human_eval}). We also provide some demo questions which will have to be correctly answered by the volunteers before we proceed with the test. We consider people with a robotics background from our research institution to volunteer for this evaluation. To reduce biases in the analysis, the person(s) responsible for designing the questions have not partaken in their evaluations. We plan to release this website in the future when \benchmark becomes public, which will also serve as a demo for our questions.

On top of the results described in Sec.~\ref{sec:evaluation}, we also perform a small scale human evaluation for the questions created from simulation environments. Specifically, we ask human volunteers to answer 10 random questions for the \emph{Place Carrot} and \emph{Sweep Object} tasks, and observe an accuracy of 100\% for both the cases. Due to the relatively smaller scale of these experiments, we do not include them as part of the discussion in the main paper.

\section{Additional Details about Real-World Experiments}
\label{appendix:real_experiments}

\begin{figure*}[t]
\begin{center}
\includegraphics[width=1.0\textwidth]{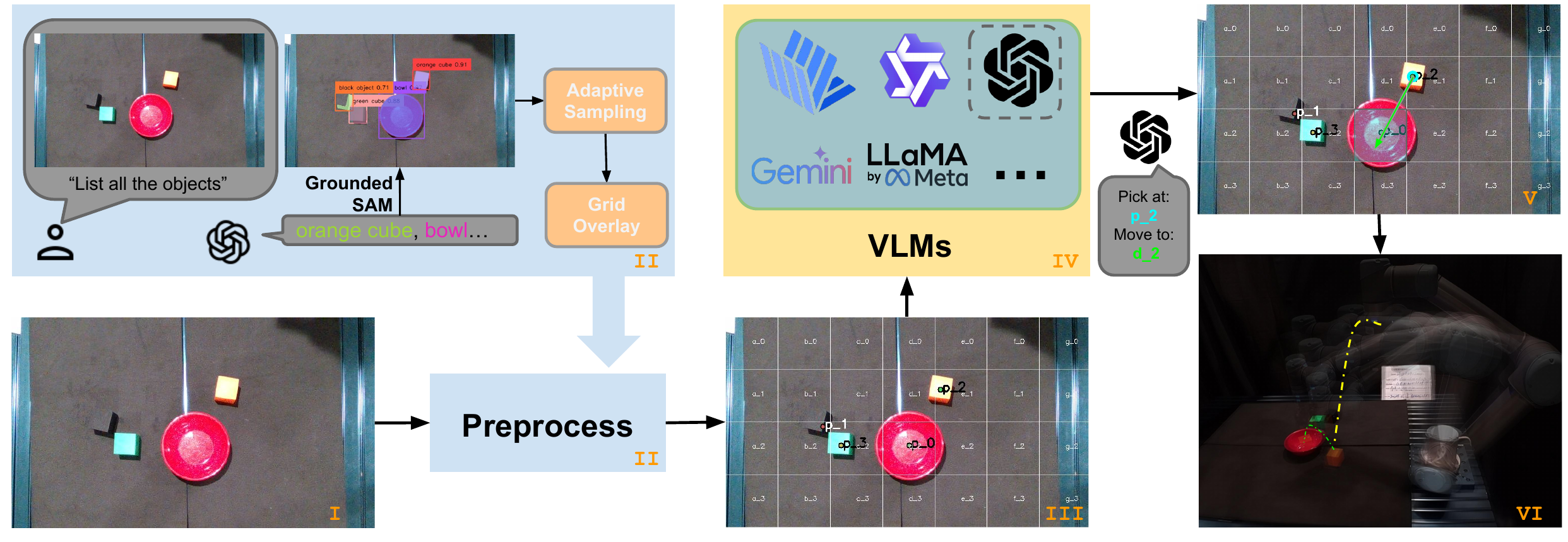}
\end{center}
\vspace{-8pt}
\caption{Real-world robot experiment pipeline. Given the task description and the top-down RGBD image (\textbf{I}), we first preprocess its RGB channels (\textbf{II}). We pass the raw RGB image to GPT and ask it to list all the objects, and then use GroundedSAM to get their masks for sampling candidate picking points. We draw the sampled points and a white grid overlay on the raw RGB image and obtain the processed image (\textbf{III}). For each task, we test different VLMs (\textbf{IV}). The chosen VLM outputs the picking point and an ending tile to complete the task. We sample a placing point inside the ending tile to ultimately get a pick-and-place action (\textbf{V}). After obtaining the pick-and-place action, and utilizing the depth information, we execute the action (\textbf{VI}).
}
\label{fig:result_real_world_appendix}
\vspace{-5pt}
\end{figure*}

We list the description of each task in the real-world experiments, which we report in Sec.~\ref{sec:evaluation}.
\begin{itemize}[noitemsep, topsep=0pt, leftmargin=*, parsep=0pt]
    \item \textbf{Color Specific p\&p}: Pick up the orange cube and place it in the bowl.
    \item \textbf{Spatial Specific p\&p}: Pick up the cube to the right of the container and place it into the container.
    \item \textbf{Fit Apple}: Pick up the apple and place it in the cardboard box that best fits it.
    \item \textbf{Pick Crumpled Cloth}: Pick up the crumpled fabric and place it in the cardboard box.
    \item \textbf{Move Fabric Easy}: Move the fabric slightly to the right while keeping it flat. To do this, lift the rightmost corner and slightly drag it to the right.
    \item \textbf{Move Fabric Intermediate}: There are two pieces of fabric. Pick up the one covering the other without disturbing the second fabric, and move it slightly to the left.
    \item \textbf{Move Fabric Hard}: Move fabric slightly to the right while keeping it flat throughout the process.
\end{itemize}

For these experiments, we introduce an adaptive point sampling scheme that samples fewer points on smaller objects, minimizing overlap among the sampled points. Additionally, we adjust the number of points sampled per object based on the task requirements to simplify tasks for VLMs. For tasks that require VLMs to reason about selecting an object among multiple options, and if the exact picking point is not critical, we sample the center point of each object's mask. 
For the MCQs, we discretize the image into a 4x7 grid to match the rectangular dimensions of the robot workspace. 
The VLMs select one grid tile as the target location for gripper placement, and the specific placement point within the selected tile is randomly sampled following a truncated multivariate Gaussian distribution with a mean at the tile center. See Fig.~\ref{fig:result_real_world_appendix} for the real-world experiment pipeline.  

\begin{figure}[htbp]
    \centering
    \begin{subfigure}[b]{0.48\textwidth}
        \centering
        \includegraphics[width=\textwidth]{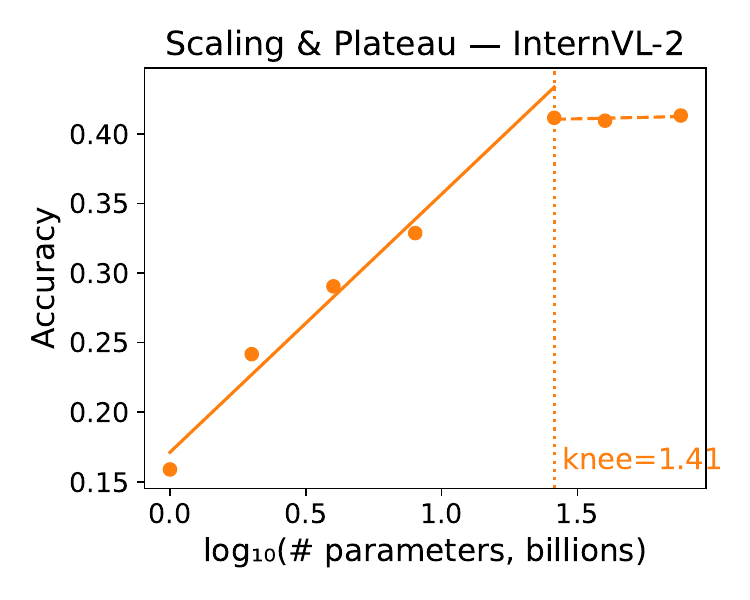}
    \end{subfigure}
    \hfill % ensures space between the two figures
    \begin{subfigure}[b]{0.48\textwidth}
        \centering
        \includegraphics[width=\textwidth]{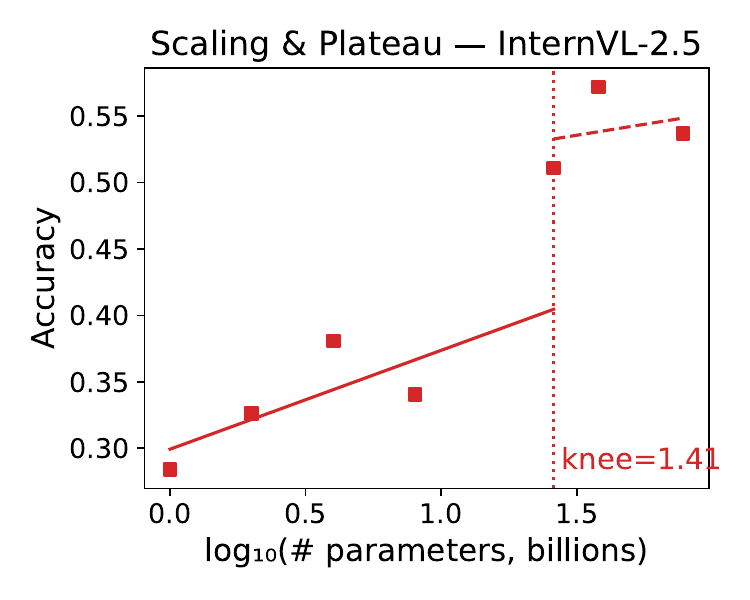}
    \end{subfigure}
    
    \caption{Log-scaling curves for the InternVL2 (left) and InternVL2.5 (right) families. Each panel plots overall model accuracy on \benchmark against $\log_{10}(\text{model size})$. A ``knee" (vertical dotted line) splits the linear growth (solid fit) regime from the post-knee plateau (dashed fit).  See App.~\ref{app:scaling_laws} for more details.}
    \label{fig:scal_plateau}
\end{figure}
\section{Additional Results from Public Robotic Manipulation Datasets }
\label{appendix:addtional_results_public}

\begin{table*}[t]
\centering

% First table: Bridge + DROID
\begin{minipage}{0.90\linewidth}
    \centering
    \setlength{\tabcolsep}{4pt}
    \scriptsize
    \begin{tabular}{l|ccc|ccc|ccc}
    \toprule
    \makecell{\bf Model} & \multicolumn{3}{c}{\bf Bridge} & \multicolumn{3}{c}{\bf DROID (art.)} & \multicolumn{3}{c}{\bf DROID (p\&p)} \\
    \midrule
    & Picking &  Ending &  Pick     &  Picking &  Ending & Pick    &  Picking &  Ending & Pick \\
    &  Point   & \ Tile   & Place    &  Point   &  Tile   & Place   &  Point   &  Tile   & Place \\
    &  Acc.    &  Acc.   & Success  &  Acc.    &  Acc.   & Success &  Acc.    &  Acc.   & Success \\

    \midrule
    \textbf{Closed-Source} & & & & & & & & &\\
    o1                     & 0.836 & 0.559 & 0.458      & \first{0.878} & 0.427 & 0.384     & \second{0.881} & \first{0.602} & \first{0.542}\\
    GPT-4.1                & 0.822 & 0.494 & 0.401      & 0.828 & 0.383 & 0.301     & \third{0.871} & 0.467 & 0.433\\
    GPT-4o                 & 0.785 & 0.395 & 0.309      & 0.843 & 0.479 & \third{0.396}     & 0.861 & 0.448 & 0.401 \\
    GPT-4o-mini            & 0.546 & 0.330 & 0.183      & 0.670 & 0.243 & 0.156     & 0.658 & 0.313 & 0.210 \\
    Gemini-2.5-pro         & \first{0.923} & 0.431 & 0.403     &\second{0.876} & 0.407 & 0.362     & \first{0.903}  &  0.493   &  0.453  \\
    % Gemini-2.0-flash-think* & ~     &  ~    & ~     &  ~    & ~     &  ~    &&&\\
    Gemini-2.0-flash       & \third{0.883} & 0.326 & 0.287      & \third{0.868} & 0.448 & 0.387          & 0.821 & 0.279 & 0.235 \\
    Gemini-1.5-pro         & 0.797 & 0.309 & 0.237      & 0.840 & 0.139 & 0.126          & 0.779 & 0.179 & 0.158 \\
    Gemini-1.5-flash       & 0.458 & 0.347 & 0.287      & 0.818 & 0.149 & 0.123          & 0.741 & 0.246 & 0.193 \\
    \midrule
    \textbf{Open-Source} & & & & & & & & &\\
    GLM-4V-9B               & 0.742 & 0.343 & 0.249     & 0.679 & \third{0.571} & 0.381     & 0.619 & 0.466 & 0.292 \\
    
    InternVL2-1B            & 0.249 & 0.135 & 0.028     & 0.262 & 0.428 & 0.105     & 0.261 & 0.254 & 0.066 \\
    InternVL2-2B            & 0.327 & 0.151 & 0.052     & 0.361 & 0.363 & 0.120     & 0.291 & 0.201 & 0.052 \\
    InternVL2-4B            & 0.445 & 0.317 & 0.134     & 0.481 & \second{0.576} & 0.357     & 0.549 & 0.434 & 0.242 \\
    InternVL2-8B            & 0.645 & 0.403 & 0.258     & 0.642 & 0.357 & 0.228     & 0.651 & 0.390 & 0.258 \\
    InternVL2-26B           & 0.725 & 0.423 & 0.326     & 0.765 & 0.379 & 0.293     & 0.759 & 0.397 & 0.296 \\
    InternVL2-40B           & 0.756 & 0.446 & 0.338     & 0.778 & 0.358 & 0.272     & 0.751 & 0.412 & 0.313 \\
    InternVL2-76B           & 0.879 & 0.588 & \third{0.509}     & 0.849 & 0.279 & 0.235     & 0.820 & 0.477 & 0.411 \\

    InternVL2.5-1B          & 0.347 & 0.035 & 0.016     & 0.379 & 0.252 & 0.099    & 0.387 & 0.165 & 0.057 \\
    InternVL2.5-2B          & 0.475 & 0.045 & 0.026     & 0.457 & 0.182 & 0.096    & 0.461 & 0.056 & 0.027 \\
    InternVL2.5-4B          & 0.681 & 0.290 & 0.199     & 0.650 & 0.280 & 0.181    & 0.549 & 0.434 & 0.242 \\
    InternVL2.5-8B          & 0.713 & 0.462 & 0.330     & 0.687 & 0.307 & 0.229    & 0.728 & 0.355 & 0.272 \\
    InternVL2.5-26B         & \second{0.884} & 0.423 & 0.376     & 0.817 & 0.527 & \first{0.433}    & 0.838 & 0.458 & 0.389 \\
    InternVL2.5-38B         & 0.875 & \second{0.606} & \second{0.528}     & 0.838 & 0.373 & 0.316    & 0.823 & 0.503 & 0.425 \\
    InternVL2.5-78B         & 0.876 & \first{0.624} & \first{0.541}     & 0.823 & 0.418 & 0.348    & 0.840 & 0.500 &0.431 \\
    
    QwenVL-Chat             & 0.291 & 0.265 & 0.067     & 0.309 & 0.326 & 0.141    & 0.259 & 0.302 & 0.077 \\
    Qwen2VL-2B              & 0.310 & 0.171 & 0.045     & 0.404 & 0.411 & 0.156    & 0.372 & 0.216 & 0.089 \\
    Qwen2VL-7B              & 0.591 & 0.329 & 0.191     & 0.611 & \first{0.593} & 0.366    & 0.650 & 0.504 & 0.329 \\
    Qwen2VL-72B             & 0.802 & 0.505 & 0.398     & 0.826 & 0.473 & 0.395    & 0.845 & 0.527 & \second{0.460} \\

    Qwen2.5-VL-3B           & 0.469 & 0.414 & 0.190    & 0.527 & 0.265 & 0.144     & 0.657 & 0.298 & 0.197 \\
    Qwen2.5-VL-7B           & 0.684 & 0.500 & 0.344    & 0.712 & 0.517 & 0.372     & 0.774 & 0.508 & 0.408 \\
    Qwen2.5-VL-32B          & 0.795 & 0.540 & 0.428    & 0.782 & 0.484 & \second{0.399}     & 0.788 & \third{0.555} & 0.459\\
    Qwen2.5-VL-72B          & 0.811 & \third{0.595} & 0.470    & 0.809 & 0.462 & 0.390     & 0.830 & \second{0.561} & \second{0.481} \\

    LLaVA-NeXT-7B           & 0.258 & 0.243 & 0.060     & 0.245 & 0.359 & 0.094    & 0.300 & 0.305 & 0.100 \\
    Llama3.2-11B-VI         & 0.388 & 0.298 & 0.118     & 0.429 & 0.276 & 0.115    & 0.413 & 0.270 & 0.111 \\
    \midrule
    Random & 0.247 & 0.279 &0.061 & 0.257 & 0.287 &0.063 & 0.250 & 0.287 & 0.084 \\
    % Human & 0.xxx & & 0.825 & 0.xxx & &0.940 & 0.xxx & &0.635 \\
    \bottomrule
    \end{tabular}
    \caption{Performance comparison of various VLMs on our MCQs for the Bridge and DROID datasets on Question Type 2 (Q2). We report the Pick Place Success as the Q2 accuracy in Table~\ref{tab:bridge_droid_results}. Details of Q2 questions can be found in Section~\ref{appendix:addtional_results_public}.}
    \label{tab:bridge_droid_addtional_results}
\end{minipage}%
\hfill
\end{table*}

We provide the MCQ answering accuracy for the questions in Table~\ref{tab:bridge_droid_results} formed from the existing datasets. Each Q1 question is an MCQ with 4 options and each Q2 question has two MCQ sub-questions and is scored as correct if the VLM answers both of them correctly:
\begin{enumerate}
    \item Picking Point Prediction: Select a pick point from the candidate picking points.
    \item Ending Tile Prediction: Given the ground-truth pick point and the task description, select the destination tile.
\end{enumerate}

We report the overall pick-place success rate as Q2 accuracy in Table~\ref{tab:bridge_droid_results}. Due to the design of Q2, we also report the picking point prediction accuracies, ending tile prediction accuracies, and overall pick-place success rates from Q2 for a more detailed comparison of various VLMs. See Table~\ref{tab:bridge_droid_addtional_results}. We use red to represent the highest accuracy, orange for the second‑highest, and yellow for the third‑highest. The same color code is used in the previous tables as well.

For the second sub-question of Q2 question, distractor tiles are sampled uniformly from all incorrect tiles, so duplicates can appear, and fewer than four options may be shown. Because the option set is random, the theoretical random performance for Q2 cannot be calculated; we instead estimate it empirically with a random policy and report it in Table~\ref{tab:bridge_droid_results} and Table~\ref{tab:bridge_droid_addtional_results}.

\noindent\textbf{Error Mode Analysis.}%\hejia{To complete, if we want.}

From Table~\ref{tab:bridge_droid_addtional_results}, the Ending Tile Accuracy is significantly lower than the Picking Point Accuracy. Our argument is that placing the object at the target location requires a more advanced and comprehensive reasoning ability of manipulation, which includes spatial reasoning, object identification, and even implicit understanding, whereas choosing the correct picking point in the questions from public robotic manipulation datasets relies more on object identification.

\noindent\textbf{Question Length Analysis.}

Our question/answer pairs are very long and they require the model to associate between the text and the image many times. To investigate, we create a small sample of 100 binary multiple choice questions to simplify the stated text-image association. These questions are designed from the Bridge~\cite{walke2023bridgedata} dataset and have the images annotated with two differently colored points (red and blue), with one of the points being a correct answer for the question under consideration. 

\begin{figure}[t]
\center
\includegraphics[width=0.8\textwidth]{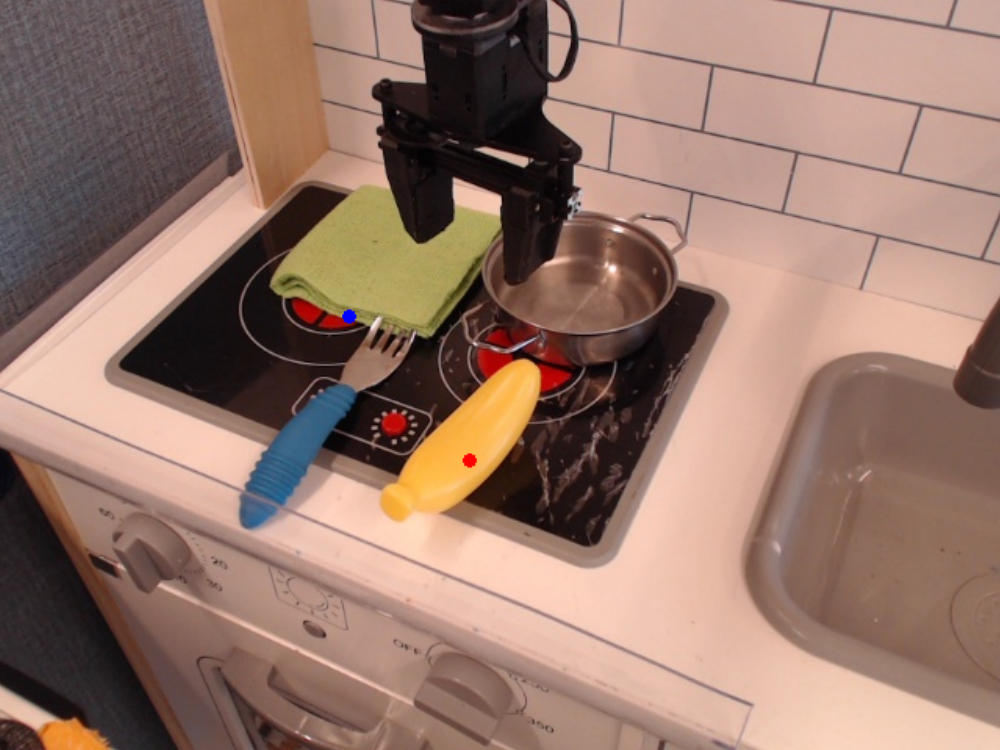}
\caption{
    Sample question for performing question length analysis. The task here is ``Put the banana on the green cloth," and the correct choice would be the red point.
}
\label{fig:sample-mcq-length}
\end{figure}

Each question has two sub-questions with different color palettes for the points. In sub-question 1, the red point is the correct picking point; In sub-question 2, the blue point is the correct picking point. The position of those 2 candidate picking points are identical on both sub-questions. This is done to mitigate possible color bias exhibited by the VLMs while answering these questions. See Fig.~\ref{fig:sample-mcq-length} for a sample question, and Table~\ref{tab:additional_results_mcq_length_analysis} for the results. We see that the results with these questions are consistent with the results stated in Section~\ref{sec:results}. For example, the best VLM is
still Gemini-2.5-pro with the overall accuracy of 0.98, while the smaller open-source VLMs demonstrate color bias and thus have relatively poor performance overall.

We also include models of the PaliGemma family~\citep{beyer2024paligemma, steiner2024paligemma, team2025gemma} in this smaller experimental study. Specifically, the original PaliGemma model with 3 billion parameters (Paligemma-V1-3B), the PaliGemma V2 model with 3 billion parameters (Paligemma-V2-3B), and the Gemma 3 model with 4 billion parameters (Gemma-3-4B). We observe that Gemma 3 performs similar to open-source models of similar size, whereas PaliGemma V2 refuses to answer the questions, resulting in a 0.00 accuracy. Since these models are not evaluated on any other questions in \benchmark, we do not include them in Table~\ref{tab:num-models}.

\begin{table*}[t]
\centering

% First table: Bridge + DROID
\begin{minipage}{\linewidth}
    \centering
    \setlength{\tabcolsep}{2pt}
    \scriptsize
    \begin{tabular}{l|c|c|c}
    \toprule
    \bf Model & \bf Accuracy & \bf Accuracy &  \bf Accuracy \\
     &  \bf Red Correct   & \bf Blue Correct   & \bf Both Correct \\    
    
    \midrule
    \textbf{Closed-Source} & & & \\
    Gemini-2.5-pro & 1.00 & 0.98 & 0.98 \\
    GPT-4o  &  0.60 &	{0.76} &	0.50 \\
    
    \midrule
    \textbf{Open-Source} & & & \\
    
    InternVL2-1B            &  0.34 &	0.20 &	0.08 \\
    InternVL2-2B           & 1.00	& 0.00	& 0.00 \\
    InternVL2-4B            &  0.12	& 1.00	& 0.12 \\
    InternVL2-8B            & 0.22 &	0.82 &	0.12 \\
    InternVL2-26B           &  0.56 &	0.66	& 0.26	\\

    InternVL2.5-1B          &  0.92 & 	0.00 &	0.00 \\
    InternVL2.5-2B          & 1.00 & 	0.12 & 	0.12 \\
    InternVL2.5-4B          & 0.10 &	0.98 &	0.10 \\
    InternVL2.5-8B          & 0.48 &	0.74 &	0.28 \\
    InternVL2.5-26B         & 0.72 &	0.76 &	0.48 \\
    
    Qwen2VL-2B              & 1.00 & 0.12 & 0.12 \\
    Qwen2VL-7B              & 0.98 &	0.58 & 	0.56 \\

    Qwen2.5-VL-3B           & 0.31 &	0.82 &	0.22 \\
    Qwen2.5-VL-7B           & 0.51 &	0.94 &	0.49 \\

    Paligemma-V1-3B          & 0.76 &	0.36 &	0.26 \\
    Paligemma-V2-3B           & 0.00 &	0.00 &	0.00 \\
    Gemma-3-4B           & 0.88 &	0.37 &	0.32 \\

    \midrule
    Random & 0.50 & 0.50 & 0.25 \\
    \bottomrule
    \end{tabular}
    \caption{Performance comparison of various VLMs on the simpler binary MCQs, which comprise of red and blue points annotated on the image with one of them being the correct answer for the task in hand. We report the question answering accuracies for the case where the red point is the correct answer, and where the blue point is the correct answer in the first two columns respectively. We also take an intersection of these and report the accuracies of solving both the sub-questions correctly in the third column.}
    \label{tab:additional_results_mcq_length_analysis}
\end{minipage}%
\hfill
\end{table*}

\begin{comment}
\section{Additional Results from Public Robotic Manipulation Datasets }
\label{appendix:addtional_results}
\end{comment}

\begin{table*}[t]
\centering

% First table: Bridge + DROID
\begin{minipage}{\linewidth}
    \centering
    \setlength{\tabcolsep}{2pt}
    \scriptsize
    \begin{tabular}{l|c|c|c|c|c|c|c|c|c|c}
    \toprule
    \bf Model & \bf High Level & \bf Fabric &  \bf Spatial     &  \bf Keypoint &  \bf Temporal & \bf Action    &  \bf Inverse &  \bf Fabric-Solid & \bf Fabric-Fabric & \bf Counter\\
     &  \bf Planning   & \bf State   & \bf Reasoning    & \bf Mapping   & \bf Sequence   & \bf Length   &  \bf Dynamics   &  \bf Interaction   & \bf Interaction & \bf Factual\\
    % &  Understanding    &  Understanding   & Abilities  &  Abilities    &  Action Sequence   & Understanding &  Understanding    &  Understanding   & Understanding & Reasoning\\

    \midrule
    \textbf{Closed-Source} & & & & & & & & & & \\
    o1  &   \second{0.879}	& \second{0.713}	& \second{0.886} &	\second{0.855}	& 0.812	& \third{0.650}	& 0.487	& \second{0.712}	& \first{0.625}	& \second{0.799} \\
    GPT-4.1  & \third{0.825}	& 0.589 &	0.587	& 0.551	& 0.745	& 0.512	& 0.291	& 0.347	& 0.282	& 0.565\\
    GPT-4o  &  0.770 &	\third{0.624} &	0.606 &	0.558 &	\third{0.829} &	\first{0.754} &	0.395 & 0.347 &	0.243 &	\third{0.632}\\
    GPT-4o-mini            & 0.662 &	0.436 &	0.409 &	0.336 &	0.537	& 0.387 &	0.358 &	0.184 &	0.171 & 0.469 \\
    Gemini-2.5-pro         & \first{0.887}	& \first{0.726} &	\first{0.975} &	\first{0.990} &	\first{0.962} &	0.487 &	\first{0.775} &	\first{0.762} &	\second{0.616} &	\first{0.820}\\
    Gemini-2.0-flash       & 0.770	& 0.556	& 0.673 &	0.750 &	\second{0.858}	& 0.533 &	0.458 &	0.298 &	\third{0.336} &	0.544 \\
    Gemini-1.5-pro         & 0.712 & 0.449 &	0.452 & 	0.445 &	0.725 &	0.475 &	0.283 &	0.230 &	0.296 & 0.527\\
    Gemini-1.5-flash       & 0.675	& 0.448 &	0.378 &	0.461 & 0.625 &	0.475 &	0.208 & 0.220 &	0.182 & 0.435\\
    \midrule
    \textbf{Open-Source} & & & & & & & & & &\\
    GLM-4V-9B               & 0.496 &	0.423 &	0.541 &	0.711 & X	& 0.216 &	X	& 0.305 &	0.146 &	0.422\\
    
    InternVL2-1B            &  0.312 &	0.167 &	0.202 &	0.313 &	0.222 &	0.333 &	0.083 &	0.327 & 0.151 & 0.192\\
    InternVL2-2B           & 0.475	& 0.372	& 0.452	& 0.426	& 0.225	& 0.221	& 0.358	& 0.326	& 0.150	& 0.293 \\
    InternVL2-4B            &  0.604	& 0.393	& 0.534	& 0.641	& 0.241	& 0.279	& 0.383	& 0.184	& 0.157	& 0.356 \\
    InternVL2-8B            & 0.612 &	0.406 &	0.455 &	0.448 &	0.354 & 0.266 &	0.446 &	0.262 &	0.186 & 0.414 \\
    InternVL2-26B           &  0.562 &	0.389	& 0.612	& 0.570	& 0.259	& 0.254	& 0.404	& 0.273	& 0.214	& 0.343\\
    InternVL2-40B           &  0.645 &	0.479 &	0.566 & 0.539 &	0.296 &	0.308 &	0.470 &	0.234 & 0.132 & 0.402\\
    InternVL2-76B           &  0.762 &	0.594	& 0.748 &	0.724 & 0.500 & 0.512 &	0.487 &	0.177 &	0.243 & 0.364\\

    InternVL2.5-1B          &  0.346 & 	0.252 &	0.295 &	0.295 &	0.304 &	0.233 &	0.316 & 0.245 &	0.232 & 0.259\\
    InternVL2.5-2B          & 0.446 & 	0.359 & 	0.517 &	0.500 &	0.150 & 	0.221 &	0.437 &	0.372 &	0.100 & 0.364\\
    InternVL2.5-4B          & 0.608 &	0.350 &	0.575 &	0.708 &	0.279 &	0.304 &	0.400 &	\third{0.549} &	0.168 &	0.368\\
    InternVL2.5-8B          & 0.658 &	0.402 &	0.504 &	0.570 &	0.433 &	0.358 &	0.471 &	0.316 &	0.182 & 0.473\\
    InternVL2.5-26B         & 0.671 &	0.589 &	0.628 &	0.574 &	0.658 &	0.354 &	0.487 &	0.284 & 0.296 & 0.439\\
    InternVL2.5-38B         & 0.758 &	0.529 &	\third{0.772} &	0.721 &	0.733 &	0.400 &	\second{0.548} & 0.213 &	0.282 & 0.594 \\
    InternVL2.5-78B         & 0.800 & 	0.487 &	0.741 &	\third{0.753} &	0.808 &	\second{0.687} &	\third{0.512} &	0.344 &	0.332 &	0.586\\
    
    QwenVL-Chat             & 0.316 &	0.295 &	0.311 &	0.311	& X	& 0.187 &	X	& 0.138 & 	0.068 &	0.209\\
    % Qwen2VL-2B              & & & & & & & & & &\\
    Qwen2VL-7B              & 0.616 &	0.389 & 	0.372 &	0.340 &	0.216 &	0.604 &	0.358 &	0.255 &	0.114 &	0.376\\
    % Qwen2VL-72B             & & & & & & & & & &\\

    Qwen2.5-VL-3B           & 0.562 &	0.364 &	0.320 &	0.317 &	0.262 &	0.204 &	0.383 &	0.294 &	0.211 &	0.343\\
    Qwen2.5-VL-7B           & 0.637 &	0.389 &	0.401 &	0.439 &	0.262 &	0.400 & 0.408 &	0.397 &	0.157 & 0.322\\
    % Qwen2.5-VL-32B          & & & & & & & & & &\\
    % Qwen2.5-VL-72B          & & & & & & & & & &\\

    \midrule
    Random & 0.237 & 0.226 & 0.240 & 0.275 & 0.246 & 0.250 & 0.267 & 0.269 & 0.246 & 0.280 \\
    Human & 0.852 & 0.960 & 0.980 & 1.000 & 1.000 & 1.000 & 0.920 & 0.910 & 0.940 & 0.742 \\
    \bottomrule
    \end{tabular}
    \caption{Performance comparison of various VLMs on our MCQs for evaluating fabric manipulation. We report accuracies for the dimensions discussed in Section~\ref{appendix:data_preparation_fabric}. For tasks requiring multi-image reasoning, the performance of models that do not support it is denoted by X in the table.}
    \label{tab:additional_results_fabric}
\end{minipage}%
\hfill
\end{table*}

\section{Additional Results from the In-house Fabric Manipulation Setup}
\label{appendix:additional_results_fabric}

We provide comparison for the MCQ answering accuracies of selected VLMs in Figure~\ref{fig:result_real_data_vedant} for the questions formed from the in-house fabric manipulation setup. The MCQ answering accuracy for all the models can be found in Table~\ref{tab:additional_results_fabric}. The models of GLM-4V-9B and QwenVL-Chat do not support multi-image reasoning hence we do not include any accuracies for these models on the tasks of \emph{Temporal Sequence Understanding} and \emph{Inverse Dynamics Understanding} as they require reasoning with two or more images.

\section{Additional Results from the Statistical Analysis}
\label{appendix:additional_results_stats}

As discussed in Section~\ref{ssec:real_world_results}, we compute the Pearson's and the Spearman's coefficients to quantify the correlation between the performance of VLMs on the questions of \benchmark and their effectiveness as real-world robotic agents. See Table~\ref{tab:stat_analysis_appendix}. On performing the analysis by aggregating over the task categories, we observe that the questions from for evaluating Fabric Manipulation have the strongest correlation with the real-world success rates with the Pearson's coefficient of 0.950 ($p=0.001$), Spearman's coefficient of 0.986 ($p=0.001$), and Kendall's Tau of 0.9090 ($p=0.002$). These values indicate an extremely strong linear relationship that maintains rank-order consistency. The questions from existing simulation data demonstrate a relatively weaker correlation, with the Pearson's coefficient being 0.779 ($p=0.023$), Spearman's coefficient being 0.638 ($p=0.009$), and Kendall's Tau being 0.691 ($p=0.018$). The reason behind this could be attributed to the weaker correlations observed for the Sweep Object task as shown in Table~\ref{tab:stat_analysis_appendix}.

\begin{table*}[t]
\centering

\begin{minipage}{0.8\linewidth}
    \centering
    \setlength{\tabcolsep}{4pt}
    \scriptsize
    \begin{tabular}{lcccccc}
    \toprule
    \makecell{\bf Task Type} & \multicolumn{2}{c}{\bf Pearson} & \multicolumn{2}{c}{\bf Spearman} & \multicolumn{2}{c}{\bf Kendall's Tau}\\
    & Coefficient & p-value & Coefficient & p-value & Coefficient & p-value \\

    \midrule
    \textbf{From Public Robotic Datasets} & & & & & & \\
    DROID pick and place (Q1) & 0.887 & 0.003 & 0.862 & 0.006 & 0.691 & 0.018 \\
    DROID articulated (Q1) & 0.905 & 0.002 & 0.898 & 0.002 & 0.764 & 0.009 \\
    Bridge (Q1) & 0.628 & 0.095 & 0.683 & 0.062 & 0.618 & 0.034 \\
    
    DROID pick and place (Q2) & 0.811 & 0.015 & 0.874 & 0.005 & 0.691 & 0.018 \\
    DROID articulated (Q2) & 0.426 & 0.293 & 0.323 & 0.435 & 0.182 & 0.533 \\
    Bridge (Q2) & 0.605 & 0.112 & 0.657 & 0.077 & 0.519 & 0.079 \\
    
    \midrule
    \textbf{For Evaluating Fabric Manipulation} & & & & \\
    
    Task Planning Understanding & 0.922 & 0.001 & 0.934 & 0.001 & 0.837 & 0.004 \\
    Fabric State Understanding & 0.905 & 0.002 & 0.934 & 0.001 & 0.837 & 0.004 \\
    Spatial Reasoning Abilities & 0.868 & 0.005 & 0.850 & 0.007 & 0.691 & 0.018 \\
    Keypoint Mapping Abilities & 0.655 & 0.078 & 0.575 & 0.136 & 0.473 & 0.105 \\
    Temporal Understanding of Action Sequence & 0.886 & 0.003 & 0.934 & 0.001 & 0.837 & 0.004 \\
    Action Length Understanding & 0.729 & 0.040 & 0.707 & 0.050 & 0.473 & 0.105 \\
    Inverse Dynamics Understanding & 0.767 & 0.026 & 0.719 & 0.045 & 0.618 & 0.034 \\
    Fabric-Solid Body Interaction Understanding & 0.788 & 0.020 & 0.850 & 0.007 & 0.764 & 0.009 \\
    Fabric-Fabric Interaction Understanding & 0.808 & 0.015 & 0.801 & 0.017 & 0.667 & 0.024 \\
    Counterfactual Understanding & 0.908 & 0.002 & 0.922 & 0.001 & 0.837 & 0.004 \\

    \midrule
    \textbf{From Existing Simulation Environments} & & & & \\
    
    Place Carrot (pick and place task) & 0.679 & 0.064 & 0.719 & 0.045 & 0.473 & 0.105 \\
    Close Drawer (articulated manipulation task) & 0.862 & 0.006 & 0.807 & 0.015 & 0.667 & 0.024 \\
    Straighten Rope (deformable manipulation task) & 0.710 & 0.048 & 0.743 & 0.035 & 0.473 & 0.105 \\
    Sweep Object (tool manipulation task) & 0.270 & 0.518 & 0.204 & 0.629 & 0.255 & 0.383 \\
    Ball Shoot (dynamic manipulation task) & 0.815 & 0.014 & 0.814 & 0.014 & 0.764 & 0.009 \\

    \bottomrule
    \end{tabular}
    \caption{Pearson’s, Spearman’s, and Kendall's Tau coefficients, along with their corresponding p-values, computed with respect to the experiments described in Section~\ref{ssec:real_world_results}.}
    \label{tab:stat_analysis_appendix}
\end{minipage}%
\hfill
\end{table*}

\section{Scaling Laws Analysis}
\label{app:scaling_laws}

Scaling laws~\cite{kaplan2020scaling} describe how a model's performance rises as more compute, data, or parameters are added. Quantifying these relationships lets practitioners decide whether a large model is worth the extra cost and guides researchers toward regimes where architectural innovation, rather than brute scale, is likely to deliver the next accuracy jump. 

We studied the scaling behaviors of 4 open-source model families: InternVL2, InternVL2.5, Qwen2VL, and Qwen2.5-VL. By computing the Pearson correlation coefficient (r), which measures how closely accuracy increases follow a straight-line relationship with $\log_{10}(\text{model size})$ we find strong scaling in every family: InternVL-2 (r=0.969), InternVL-2.5 (0.937), Qwen-VL-2 (0.998), Qwen-VL-2.5 (0.890). 

Furthermore, we also performed a local, post-knee analysis on the two 2 InternVL families (Fig.~\ref{fig:scal_plateau}) to study their scaling plateauing behaviors. Using a knee-finder that keeps at least three checkpoints on each side of the split, we fitted separate lines before and after the knee. Focusing only on the few checkpoints that lie just beyond each knee, we find that the growth slope nearly vanishes.

\section{Sample Multiple Choice Question Examples}
\label{appendix:additional_questions}

This section contains one sample question for each question type in \benchmark. We include the questions with the exact text instruction given as prompt to the VLMs during the evaluation. The figure captions are self-explanatory.

\subsection{From Public Robotic Manipulation Datasets}

\subsubsection{Type 1 (Q1)}

\begin{figure}[t]
\center
\includegraphics[width=0.8\textwidth]{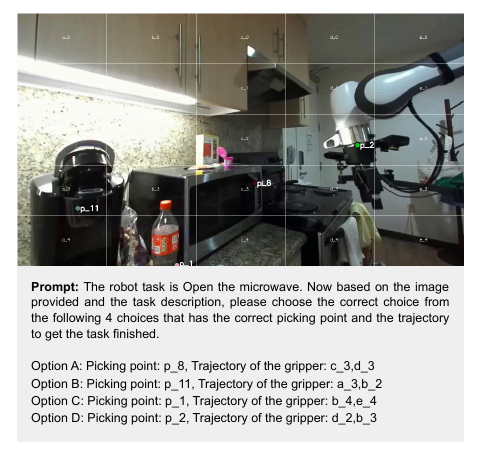}
\vspace{-10pt}
\caption{
    Sample question for \textit{Type 1 (Q1)} from Existing Robotic Manipulation Datasets. Answer: Option A. 
}
\label{fig:sample-real-q1}
\end{figure}

The left-most question in Fig.~\ref{fig:pull} is an example question in this category. Another sample question (with the exact VLM prompts) can be found in Fig.~\ref{fig:sample-real-q1}. This question corresponds to the DROID articulated (art.) task.

\subsubsection{Type 2 (Q2)}

\begin{figure}[t]
\center
\includegraphics[width=0.8\textwidth]{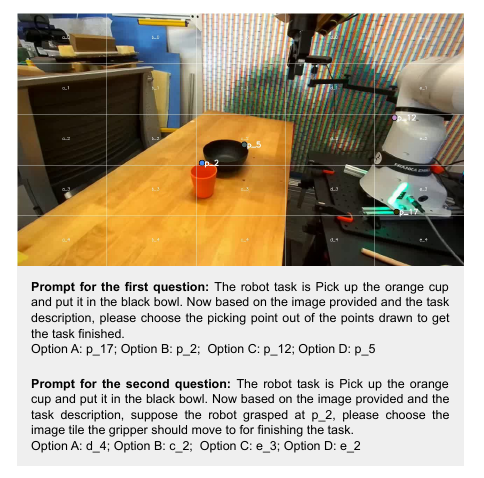}
\vspace{-10pt}
\caption{
    Question for \textit{Type 2 (Q2)} from Existing Robotic Manipulation Datasets. The answer to the first question is B and the answer to the second question is B. The VLM has to answer both questions correctly. 
}
\label{fig:sample-real-q2}
\end{figure}

A sample question with the exact VLM prompts for this question type is in Fig.~\ref{fig:sample-real-q2}. This question corresponds to the DROID pick and place (p\&p) task. 

% Each Q2 question has two sub-questions and is scored as correct if the VLM answer both of them correctly:
% \begin{enumerate}
%     \item Picking Point Prediction: Select a pick point from the candidate picking points.
%     \item Ending Tile Prediction: Given the ground-truth pick point and the task description, select the destination tile.
% \end{enumerate}

% For the second sub-question, distractor tiles are sampled uniformly from all incorrect tiles, so duplicates can appear and fewer than four options may be shown. Because the option set is random, the theoretical random performance for Q2 cannot be calculated; we instead estimate it empirically with a random policy and report it in Table~\ref{tab:bridge_droid_results} and Table~\ref{tab:bridge_droid_addtional_results}.

\subsection{From In-house Fabric Manipulation Setup}

\subsubsection{Task Planning Understanding}

\begin{figure}[t]
\center
\includegraphics[width=0.8\textwidth]{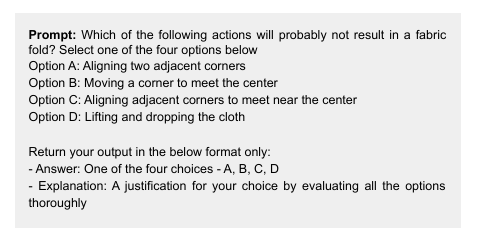}
\vspace{-10pt}
\caption{
    Question for \textit{Task Planning Understanding}. Answer: Option D.
}
\label{fig:sample-task-planning}
\end{figure}

See Fig.~\ref{fig:sample-task-planning}. 

\subsubsection{Fabric State Understanding}

\begin{figure}[t]
\center
\includegraphics[width=0.8\textwidth]{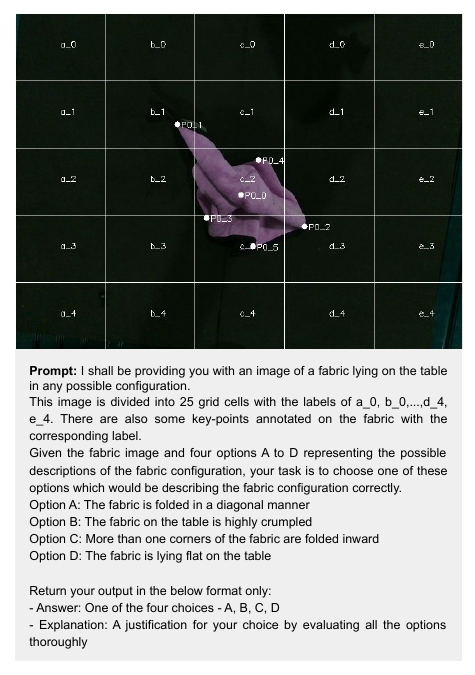}
\vspace{-10pt}
\caption{
    Question for \textit{Fabric State Understanding}. Answer: Option B.
}
\label{fig:sample-fabric-state}
\end{figure}

See Fig.~\ref{fig:sample-fabric-state}. 

\subsubsection{Spatial Reasoning Abilities}

\begin{figure}[t]
\center
\includegraphics[width=0.8\textwidth]{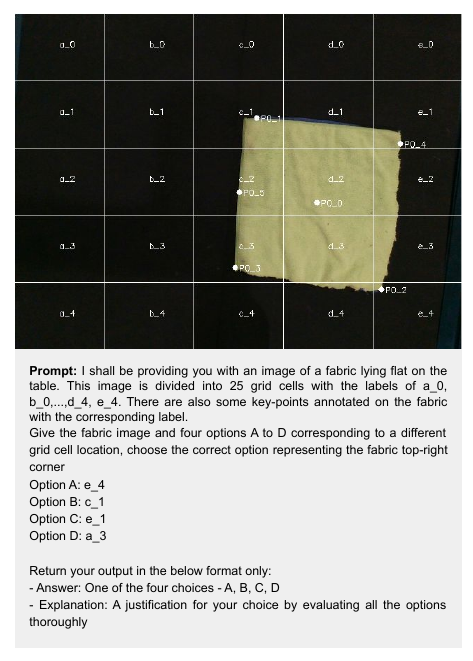}
\vspace{-10pt}
\caption{
    Question for \textit{Spatial Reasoning Abilities}. Answer: Option C.
}
\label{fig:sample-spatial-understanding}
\end{figure}

See Fig.~\ref{fig:sample-spatial-understanding}.

\subsubsection{Key-point Mapping Abilities}

\begin{figure}[t]
\center
\includegraphics[width=0.8\textwidth]{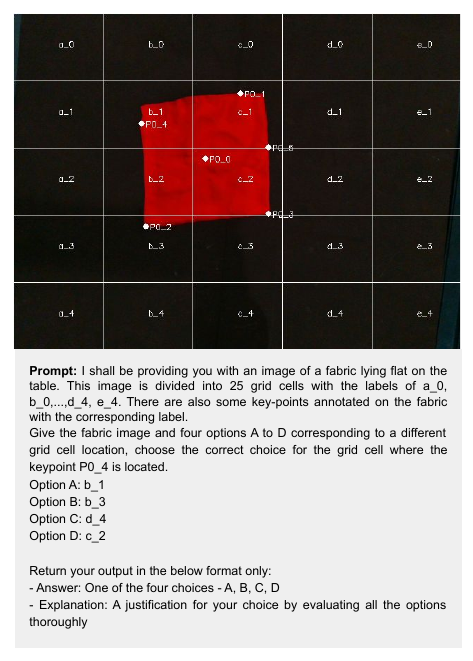}
\vspace{-10pt}
\caption{
    Question for \textit{Key-point Mapping Abilities}. Answer: Option A.
}
\label{fig:sample-keypoint-mapping}
\end{figure}

See Fig.~\ref{fig:sample-keypoint-mapping}.

\subsubsection{Temporal Understanding of Action Sequence}

\begin{figure}[t]
\center
\includegraphics[width=0.8\textwidth]{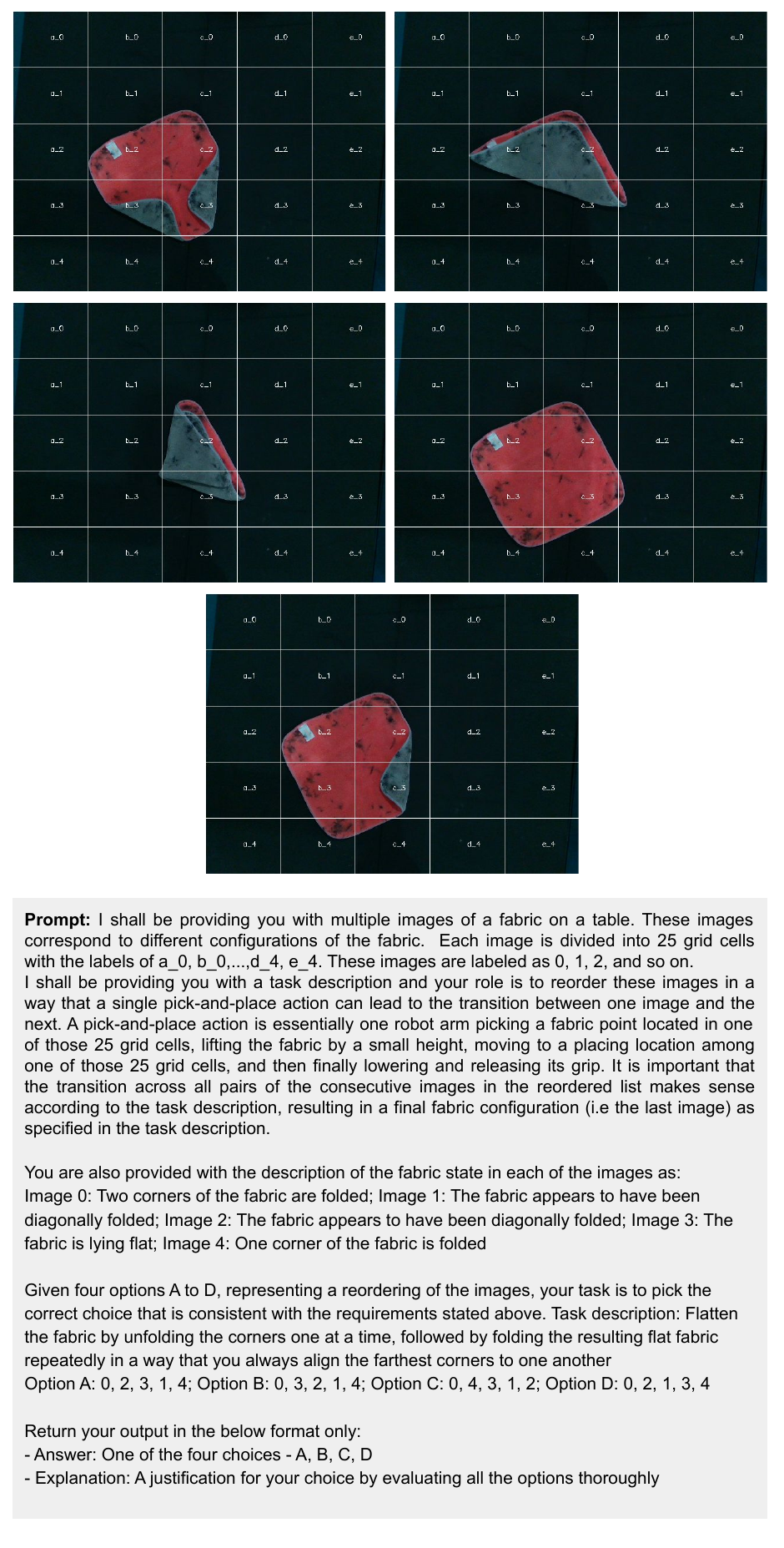}
\vspace{-10pt}
\caption{
    Sample Question for \textit{Temporal Understanding of Action Sequence}. Answer: Option C.
}
\label{fig:sample-temporal-sequence}
\end{figure}

See Fig.~\ref{fig:sample-temporal-sequence}.

\subsubsection{Action Length Understanding}

\begin{figure}[t]
\center
\includegraphics[width=0.8\textwidth]{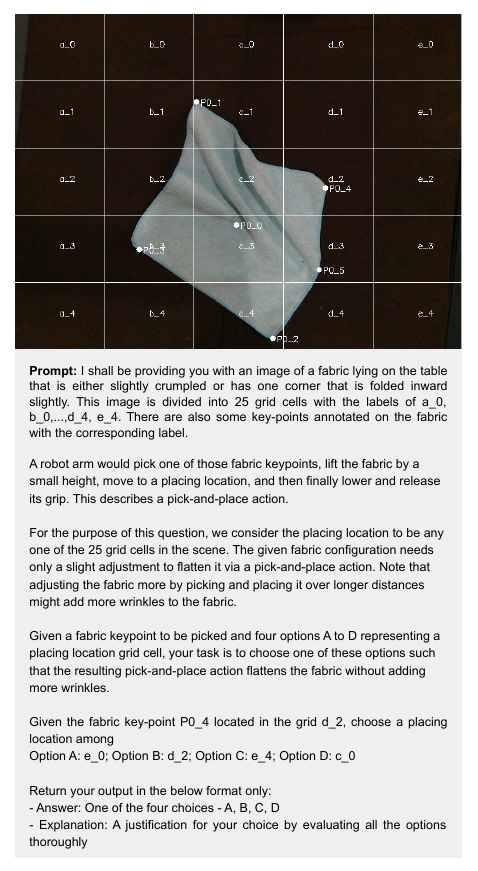}
\vspace{-10pt}
\caption{
    Question for \textit{Action Length Understanding}. Answer: Option B.
}
\label{fig:sample-action-length}
\end{figure}

See Fig.~\ref{fig:sample-action-length}.

\subsubsection{Inverse Dynamics Understanding}

\begin{figure}[t]
\center
\includegraphics[width=0.8\textwidth]{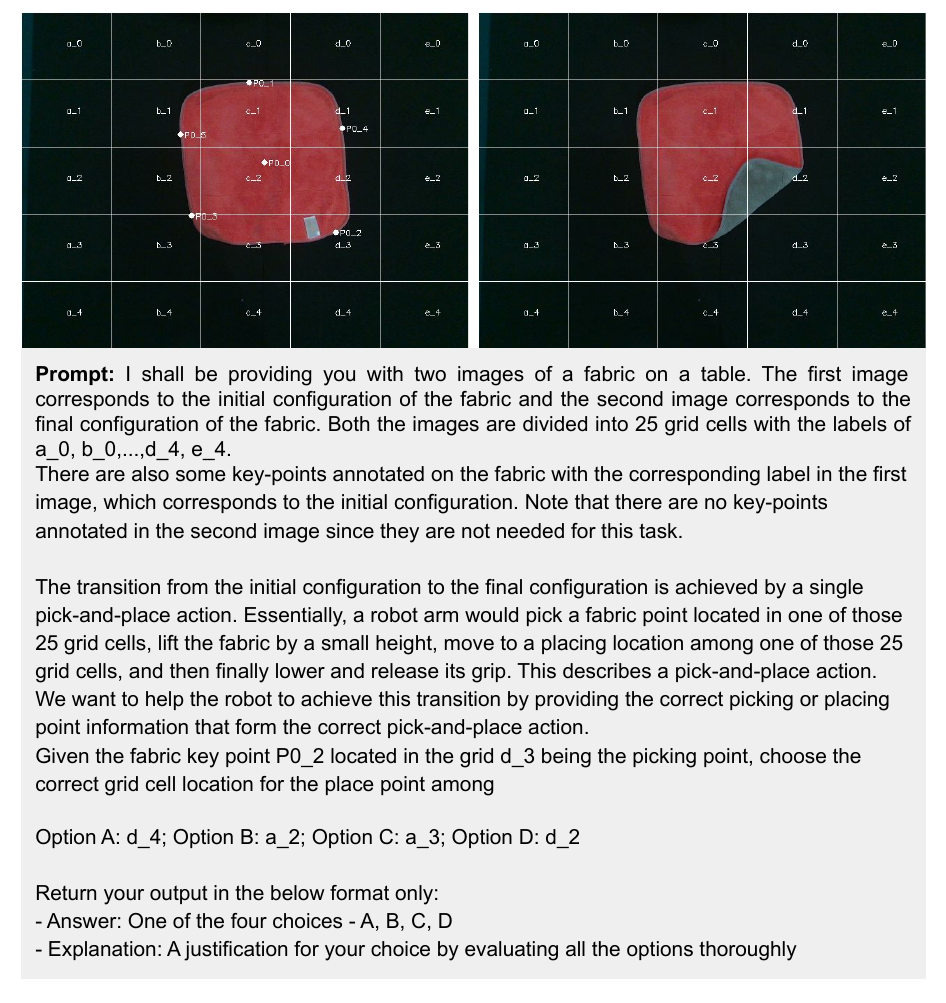}
\vspace{-10pt}
\caption{
    Question for \textit{Inverse Dynamics Understanding}. Answer: Option D.
}
\label{fig:sample-inverse-dynamics}
\end{figure}

See Fig.~\ref{fig:sample-inverse-dynamics}.

\subsubsection{Fabric-Solid Body Interaction Understanding}

\begin{figure}[t]
\center
\includegraphics[width=0.8\textwidth]{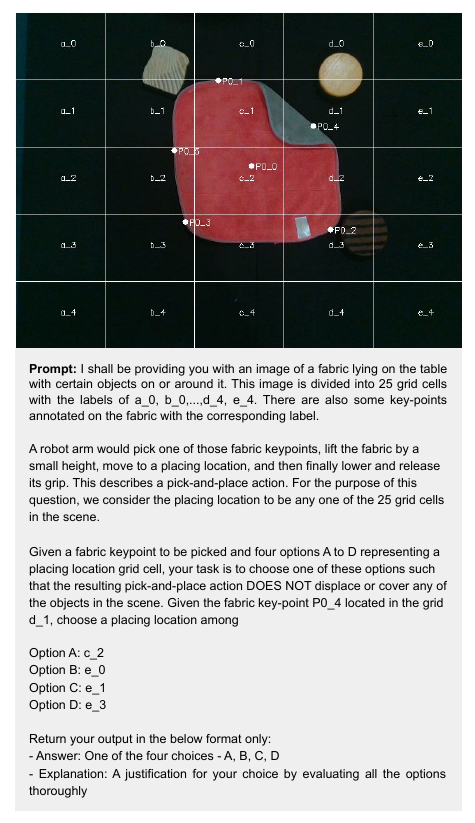}
\vspace{-10pt}
\caption{
    Sample Question for \textit{Fabric-Solid Body Interaction Understanding}. Answer: Option A.
}
\label{fig:sample-fabric-solid}
\end{figure}

See Fig.~\ref{fig:sample-fabric-solid}.

\subsubsection{Fabric-Fabric Interaction Understanding}

\begin{figure}[t]
\center
\includegraphics[width=0.8\textwidth]{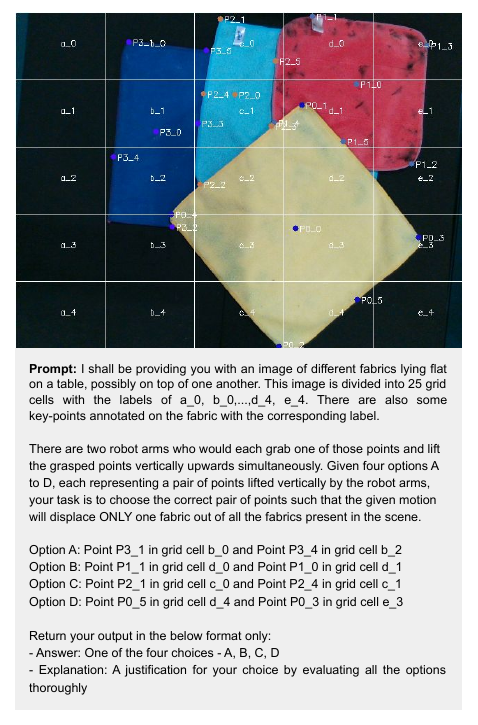}
\vspace{-10pt}
\caption{
    Sample Question for \textit{Fabric-Fabric Interaction Understanding}. Answer: Option D.
}
\label{fig:sample-fabric-fabric}
\end{figure}

See Fig.~\ref{fig:sample-fabric-fabric}.

\subsubsection{Counter Factual Understanding}

\begin{figure}[t]
\center
\includegraphics[width=0.8\textwidth]{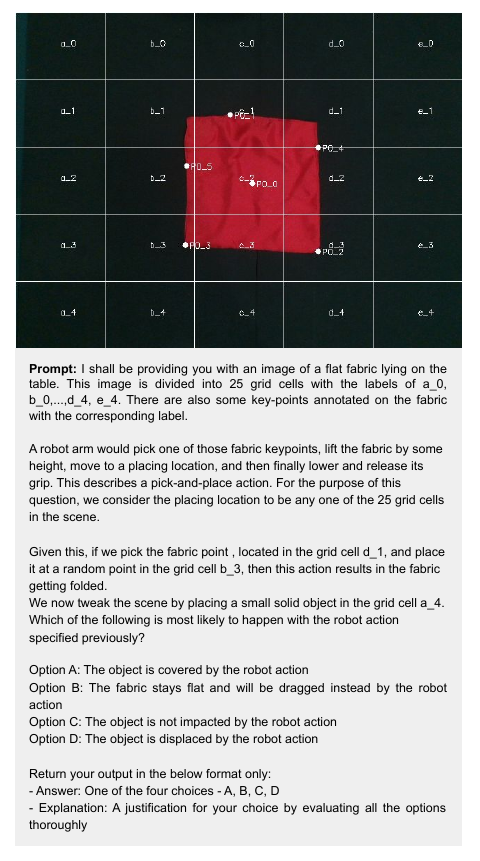}
\vspace{-10pt}
\caption{
    Question for \textit{Counter Factual Understanding}. Answer: Option C.
}
\label{fig:sample-counter-factual}
\end{figure}

See Fig.~\ref{fig:sample-counter-factual}.

\subsection{From Existing Simulation Environments}

\subsubsection{Place Carrot}

\begin{figure}[t]
\center
\includegraphics[width=0.8\textwidth]{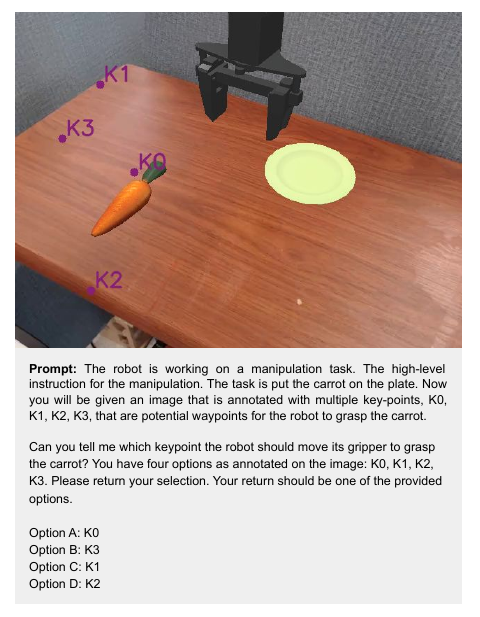}
\vspace{-10pt}
\caption{
    Question for \textit{Place Carrot} in Simulation. Answer: Option A.
}
\label{fig:sample-place-carrot}
\end{figure}

See Fig.~\ref{fig:sample-place-carrot}.

\subsubsection{Close Drawer}
\begin{figure}[t]
\center
\includegraphics[width=0.8\textwidth]{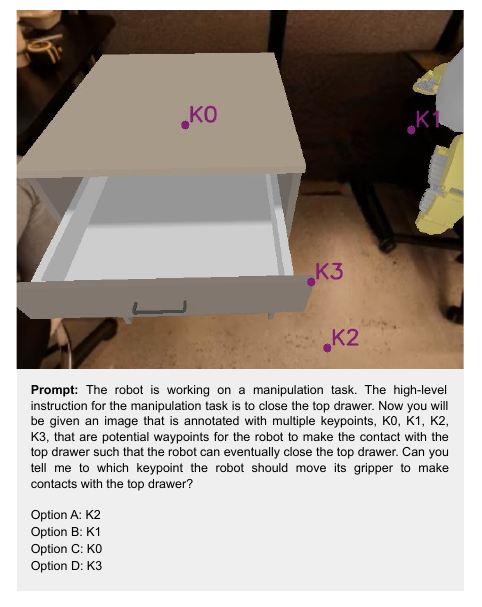}
\vspace{-10pt}
\caption{
    Question for \textit{Close Drawer} in Simulation. Answer: Option D.
}
\label{fig:sample-close-drawer}
\end{figure}

See Fig.~\ref{fig:sample-close-drawer}.

\subsubsection{Straighten Rope}
\begin{figure}[t]
\center
\includegraphics[width=0.8\textwidth]{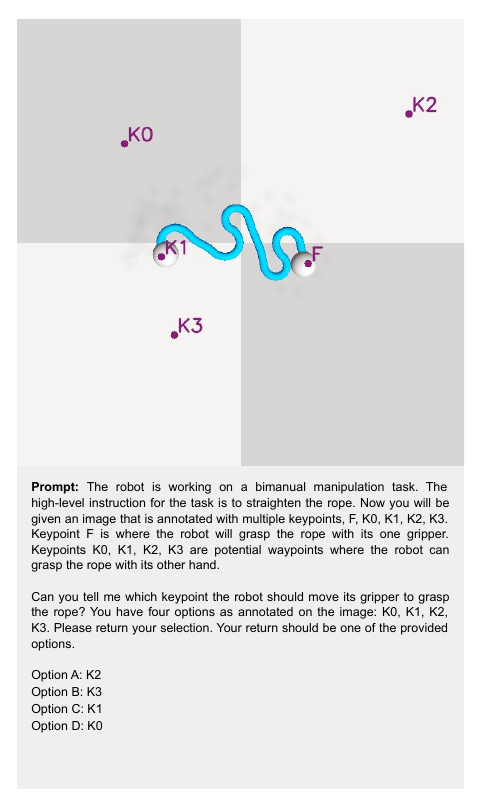}
\vspace{-10pt}
\caption{
    Question for \textit{Straighten Rope} in Simulation. Answer: Option C.
}
\label{fig:sample-straighten-rope}
\end{figure}

See Fig.~\ref{fig:sample-straighten-rope}.

\subsubsection{Sweep Object}
\begin{figure}[t]
\center
\includegraphics[width=0.8\textwidth]{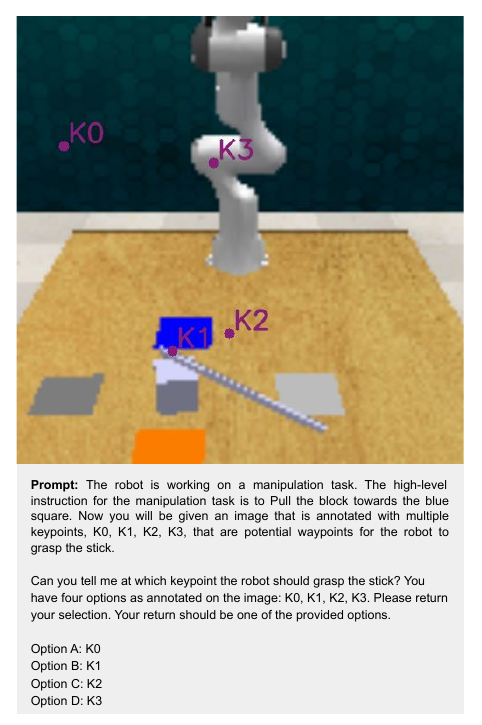}
\vspace{-10pt}
\caption{
    Question for \textit{Sweep Object} in Simulation. Answer: Option B.
}
\label{fig:sample-sweep-object}
\end{figure}

See Fig.~\ref{fig:sample-sweep-object}.

\subsubsection{Ball Shooting}\label{sec:ball_shoot}

\begin{figure}[t]
\center
\includegraphics[width=0.8\textwidth]{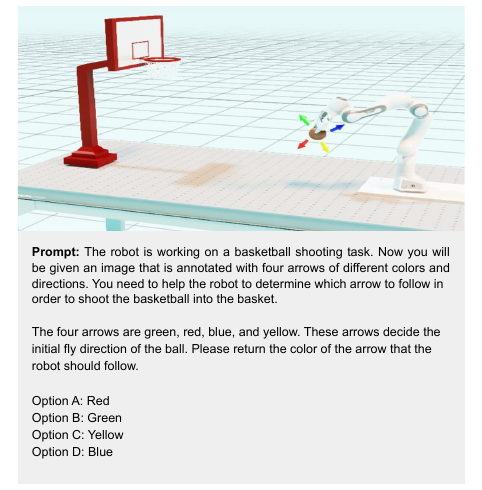}
% \vspace{-10pt}
\caption{
    Type 1 Question for the \textit{Ball Shooting} task in simulation. Correct Answer: Option B.
}
\label{fig:sample-ball-shoot-type1}
\end{figure}

\noindent\textbf{Type 1.} See Fig.~\ref{fig:sample-ball-shoot-type1}.

\begin{figure}[t]
\center
\includegraphics[width=0.8\textwidth]{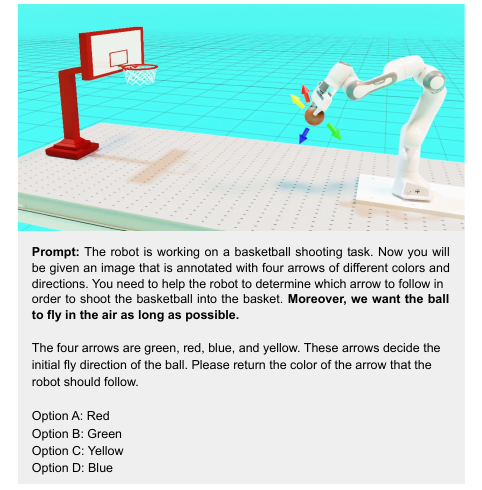}
% \vspace{-10pt}
\caption{
    Type 2 Question for the \textit{Ball Shooting} task in simulation. Correct Answer: Option A.
}
\label{fig:sample-ball-shoot-type2}
\end{figure}

\noindent\textbf{Type 2.} See Fig.~\ref{fig:sample-ball-shoot-type2}.

\begin{figure}[t]
\center
\includegraphics[width=0.8\textwidth]{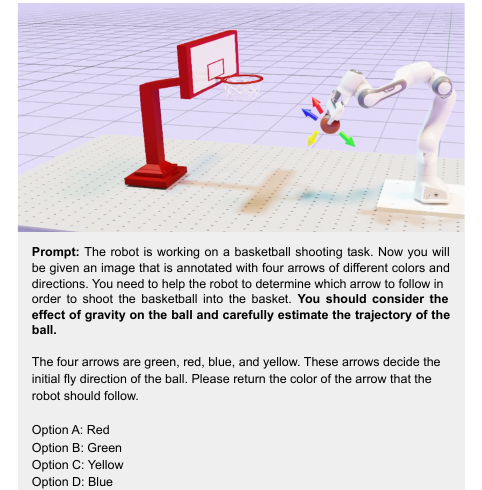}
% \vspace{-5pt}
\caption{
    Type 3 Question for the \textit{Ball Shooting} task in simulation. Correct Answer: Option A.
}
\label{fig:sample-ball-shoot-type3}
\end{figure}

\noindent\textbf{Type 3.} See Fig.~\ref{fig:sample-ball-shoot-type3}.

\end{document}

%% file: config.tex
\usepackage{array}
\usepackage{booktabs}
\usepackage{tabularx}
\usepackage[font=footnotesize, skip=5pt]{caption}

\usepackage[toc,page,header]{appendix}
\usepackage{minitoc}
\usepackage{enumitem}

\usepackage[numbers]{natbib}
\usepackage{multicol}
\usepackage{url}
\usepackage{verbatim}
\usepackage{pifont}
\usepackage{booktabs}
\usepackage{xspace}
\usepackage{graphicx}
\usepackage{colortbl}
\usepackage{xcolor}
\usepackage{ragged2e}
\usepackage{amsmath} 
\usepackage{subcaption}

\definecolor{First}{rgb}{1.00,0.75,0.75}   
\definecolor{Second}{rgb}{1.00,0.85,0.55}  
\definecolor{Third}{rgb}{1.00,0.96,0.70}   

\newcommand{\first}[1]{\cellcolor{First}{#1}}
\newcommand{\second}[1]{\cellcolor{Second}{#1}}
\newcommand{\third}[1]{\cellcolor{Third}{#1}}

\newcommand{\benchmark}{ManipBench\xspace}
\newcommand{\QuestionNum}{12617}

\newcommand{\ie}{i.e.,\xspace}

\newcolumntype{R}[1]{>{\RaggedLeft\arraybackslash}p{#1}}

\usepackage{pifont}
\newcommand{\cmark}{{\color{green}\ding{51}}}
\newcommand{\xmark}{{\color{red}\ding{55}}}  

\usepackage{makecell}